\documentclass[sigconf,natbib=false]{acmart}
\newif\ifincludeappendices
\includeappendicestrue
\AtBeginDocument{%
  }

\setcopyright{acmlicensed}
\setcopyright{cc}
\setcctype{by}
\copyrightyear{2026}
\acmYear{2026}
\acmDOI{10.1145/3806096.3844871}
\acmConference[AI Summit '26]{ACM AI Summit 2026}{August 31-September 02, 2026}{Atlanta, GA, USA}
\acmBooktitle{ACM AI Summit 2026 (AI Summit '26), August 31-September 02, 2026, Atlanta, GA, USA}
\acmISBN{979-8-4007-2638-5/2026/08}

\usepackage{booktabs}
\usepackage{enumitem}
\usepackage{placeins}
\usepackage{makecell}
\usepackage{fvextra}
\fvset{breaklines=true, breakanywhere=true, fontsize=\small}
\RequirePackage[
  datamodel=acmdatamodel,
  style=acmauthoryear,
]{biblatex}
\begin{document}

\title{Why AI Detection Fails for Academic Integrity}

\author{Jonathan A. Karr Jr.}
\authornote{These authors contributed equally to this research.}
\email{jkarr@nd.edu}

\affiliation{%
  \institution{University of Notre Dame}
  \city{Notre Dame}
  \state{Indiana}
  \country{USA}
}

\author{Grigorii Khvatskii}
\authornotemark[1]
\email{gkhvatsk@nd.edu}

\affiliation{%
  \institution{University of Notre Dame}
  \city{Notre Dame}
  \state{Indiana}
  \country{USA}
}

\author{Ting Hua}
\email{thua@nd.edu}

\affiliation{%
  \institution{University of Notre Dame}
  \city{Notre Dame}
  \state{Indiana}
  \country{USA}
}

\author{Nitesh V. Chawla}
\email{nchawla@nd.edu}

\affiliation{%
  \institution{University of Notre Dame}
  \city{Notre Dame}
  \state{Indiana}
  \country{USA}
}

\renewcommand{\shortauthors}{Karr and Khvatskii et al.}

\begin{abstract}
Institutions use commercial AI detectors for academic integrity, yet detectors cannot distinguish AI editing from full LLM drafts and may treat both as misconduct. In a controlled study of published English abstracts (four domains; 2013 to 2015 vs.\ 2023 to 2025), we quantify this policy failure under proxy human/AI labels at $\tau=0.50$. Light \emph{refine (abstract only)} edits, a proxy for guideline-compliant AI assistance, are flagged at 38 to 80\%. Unmodified 2023 to 2025 originals are flagged at 9 to 15\%, with non-STEM rates far above STEM ($p<0.001$); elevated scores track long-token and Academic Word List density, not authorship intent alone. After Undetectable AI humanization, evasion is near-total: fewer than 4\% of AI-labeled rewrites remain flagged (post-humanization detection rate $<4\%$; FNR $>96\%$). Honest AI-editing results in a higher sanction risk than humanizer-assisted evasion. Therefore, detector scores should not serve as standalone misconduct evidence.
\end{abstract}




\maketitle
\begingroup
\renewcommand{\thefootnote}{}
\footnotetext{Code: \url{https://github.com/JonathanKarr33/ai_humanization_and_detection_evaluation}}
\endgroup

\section{Introduction}
Consider two students submitting an essay or research paper. One author creates the claims, but uses LLMs to refine their language. The other student uses the LLM to produce the full draft with no original analysis. At $\tau=0.50$, commercial detectors often flag both at high rates. Yet they reflect different relationships to the work. \emph{Do they deserve the same outcome?}

LLMs have raised concerns about synthetic text in academia \cite{kasneci2023chatgpt,cotton2024chatting,perkins2023academic}, and institutions increasingly deploy AI detectors \cite{yan2023detection,wang2023synthetic}. Vendors report high benchmark accuracy \cite{spero2025pangram_ai_assistance,adam2026gptzero}, but independent studies find substantial false positives and false negatives \cite{dik2025assessing,elazar2026llm}. 

Those errors are not limited to texts possibly contaminated with LLM outputs. On pre-LLM-era (2013--2015) \emph{original} abstracts, Pangram and GPTZero incur 0\% proxy FPR at $\tau=0.50$ (Table~\ref{tab:error_rates}). On 2023--2025 originals, the same threshold yields 14.9\% (Pangram) and 8.9\% (GPTZero) \emph{flag rates}(not confirmed false positives, because unobserved AI assistance is possible) with non-STEM rates far above STEM ($p<0.001$). Detectors therefore behave unevenly across scholarly domains even on peer-reviewed abstract prose.

Detectors do not separate fully AI-generated from AI-assisted prose in one score \cite{spero2025pangram_ai_assistance,gptzero2026benchmarking,pratama2025accuracy}. At the same time, the growing market for humanizers, software designed to evade AI detectors, adds another potential source for detection errors \cite{wallwork2025ai}. Furthermore, English-language learners may legitimately use AI for translation, clarity, or academic polish while retaining intellectual ownership \cite{zhang2023don}, yet such edits can elevate detector scores.  These facts together create a tension where people using AI in legitimate ways to present their ideas are punished by AI detection pipelines, but people willing to cheat by using AI in combination with a humanizer are not. To analyze this tension, we ask three policy-oriented research questions focused on \emph{detection errors}, not vendor benchmarking:

\begin{figure*}
    \centering
    \includegraphics[width=1\linewidth]{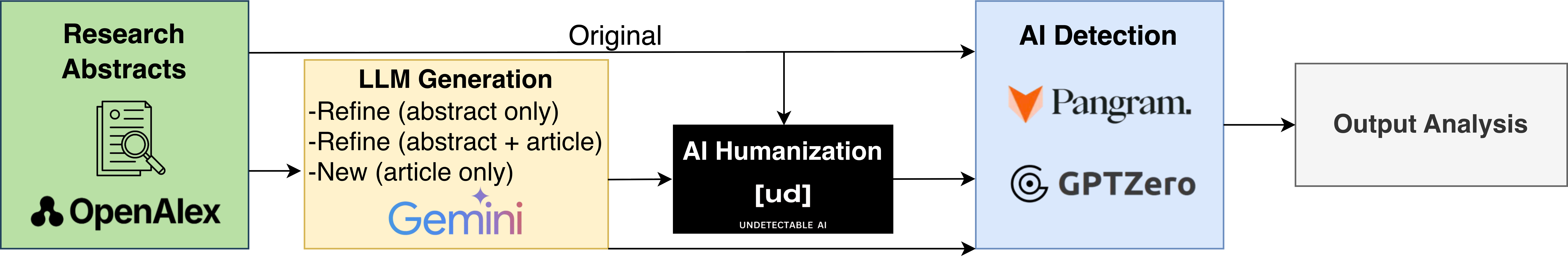}
    \caption{AI Detection Pipeline}
    \label{ai_detection_pipeline}
\Description{Pipeline diagram connecting OpenAlex research abstracts, three Gemini rewrite conditions, Undetectable AI humanization, Pangram and GPTZero detection, and output analysis. The rewrite conditions use the abstract alone, the abstract with the article, or the article alone.}
\end{figure*}

\begin{description}[style=nextline]
\item[\textbf{RQ1}] Under proxy ground-truth labels, what are the false-positive and false-negative rates of commercial detectors, and how do they vary by domain, time period, and rewrite condition?

\item[\textbf{RQ2}] Which surface linguistic features of academic abstracts are associated with elevated detector scores (and thus higher false-positive risk on human writing)?

\item[\textbf{RQ3}] How does humanization change false-negative rates on AI-generated text?
\end{description}

Prior work has documented detector unreliability \cite{dik2025assessing}, including cases in which polishing abstracts are misclassified as full rewrites \cite{geng2025we}. We extend this literature with a unified error-analysis framework: proxy-labeled positives/negatives, permutation-based inference, and interpretable text-feature correlates across domains and time windows. Findings are intended to inform policy on published abstract proxies.

\section{Methodology}

We describe our corpus and detection pipeline and define how we operationalize detector error rates and the accompanying statistical analyses. In short, we score original and LLM-rewritten abstracts across four domains and two time periods with two commercial detectors, before and after humanization.

\subsection{Corpus and Pipeline}
\label{sec:corpus_pipeline}
We sampled English abstracts from OpenAlex \cite{priem2022openalex} in chemistry, computer science, political science, and theology (100 per domain-time bucket before filtering) for 2013 to 2015 and 2023 to 2025. The earlier window provides a pre-LLM-era reference for proxy FPR on \emph{original} abstracts; the later window reports flag rates when LLM-assisted writing is plausible but unobserved in our labels \cite{liang2024monitoring,kobak2025delving,geng2025we}. After requiring 25 to 500 words and PDF access ($\approx$20\% loss), we retained 642 abstracts. Each contributed an \emph{original} abstract plus three Gemini 3 Flash rewrites via OpenRouter \cite{openrouter_ai}: \emph{refine (abstract only)}, \emph{refine (abstract + article)}, and \emph{new (article only)}. We scored all variants with Pangram 3.2 and GPTZero \cite{emi2024technical,adam2026gptzero} ($s \in [0,1]$) and an LLM-assisted baseline (\texttt{GPT-5 Nano}). We then humanized all variants with Undetectable AI v11 (Balanced/Doctorate/Article) and re-scored outputs (Fig.~\ref{ai_detection_pipeline}).

STEM versus non-STEM tests pool chemistry with computer science and political science with theology. Table~\ref{tab:corpus_counts} lists retained abstracts per domain-time bucket. The three rewrite conditions hold the generator fixed: \emph{refine (abstract only)} receives only the abstract, \emph{refine (abstract + article)} also receives the article for context, and \emph{new (article only)} writes a fresh abstract from the article body. Both refine prompts request rewriting to improve the prose and make it sound more human-like, with at least some changes. They do not restrict edits to grammar correction. The new-abstract prompt requests natural prose without copying the article verbatim. Pangram and GPTZero use the same threshold, $\tau=0.50$ (sensitivity in Table~\ref{tab:threshold_sensitivity}).

\begin{table}[t]
\centering
\small
\begin{tabular}{@{}lrr@{}}
\toprule
Domain & 2013--15 & 2023--25 \\
\midrule
Chemistry & 84 & 95 \\
Computer science & 70 & 83 \\
Political science & 78 & 79 \\
Theology & 74 & 79 \\
\bottomrule
\end{tabular}
\caption{Retained abstracts by domain after filtering ($N=642$). Pangram and GPTZero cover all papers.}
\label{tab:corpus_counts}
\Description{Counts of retained abstracts in chemistry, computer science, political science, and theology for 2013--2015 and 2023--2025. The two periods contain 306 and 336 abstracts, respectively, for a total of 642.}
\end{table}

\subsection{Error Rates and Analysis}
\label{sec:error_operationalization}
With $\tau=0.50$ and $\hat{y}_i=\mathbb{I}\{s_i\ge\tau\}$, we assign proxy human labels to originals and proxy AI labels to LLM outputs, defining $\mathrm{FP}_i=\mathbb{I}\{y_i=\text{human}\land\hat{y}_i=1\}$ and $\mathrm{FN}_i=\mathbb{I}\{y_i=\text{AI}\land\hat{y}_i=0\}$. We report FNR on LLM-labeled variants and proxy FPR on 2013 to 2015 \emph{original} abstracts. For 2023 to 2025 originals, we report \emph{flag rate}, because unobserved AI assistance prevents interpreting positives as confirmed false positives.

We also report \emph{AI-assisted false-positive risk}: the fraction of \emph{refine (abstract only)} outputs flagged under proxy human-source labels (Table~\ref{tab:assisted_fp_risk}). This assisted-writing flag rate equals $1-\mathrm{FNR}$ under proxy AI labels for the same texts; it is not FPR on unmodified human writing. Whether this assistance is permitted depends on institutional rules. Our prompt models generative rewriting, not verified grammar-only assistance. The policy comparison with humanized rewrites therefore spans different interpretations of authorship, rather than one misconduct confusion matrix.

Group comparisons use two-sided permutation tests on mean score differences (5,000 iterations; seed 42). Long-token ratio counts tokens with at least 10 characters. Academic Word List (AWL) density matches normalized tokens to Coxhead's 570 headwords by Porter stemming \cite{coxhead2000new}; type-token ratio divides unique case-folded tokens by all tokens. Numeric and non-alphabetic shares count tokens containing digits and non-letters, respectively; acronym density counts uppercase tokens of at least two characters. We report Spearman $\rho$ with Benjamini-Hochberg FDR within each collection and domain-centered associations. Length-adjusted correlations residualize features and scores on $\log(1+\text{word count})$; word count uses whitespace splitting.

\section{Results}

We organize our findings around the three research questions, reporting proxy-labeled error rates (\textbf{RQ1}), the linguistic correlates of detector scores (\textbf{RQ2}), and the effects of humanization on detector behavior and agreement (\textbf{RQ3}). Across all three, detectors prove sensitive to stylistic cues and are readily defeated by humanizers, leading to high FNR and low inter-detector agreement.

\begin{table}[t]
  \centering
  \small
  \setlength{\tabcolsep}{3pt}
  \begin{tabular}{@{}llrrr@{}}
  \toprule
  Period & Detector & FPR/flag & FNR pre & FNR post \\
  \midrule
  2013-15 & Pangram & 0.0\% & 19.4\% & 98.5\% \\
  2013-15 & GPTZero & 0.0\% & 44.2\% & 96.0\% \\
  2013-15 & LLM-aid & 46.7\% & 35.1\% & 30.5\% \\
  2023-25 & Pangram & 14.9\% & 13.5\% & 96.6\% \\
  2023-25 & GPTZero & 8.9\% & 38.4\% & 96.1\% \\
  2023-25 & LLM-aid & 70.8\% & 22.8\% & 19.7\% \\
  \bottomrule
  \end{tabular}
 \caption{Proxy-labeled error rates at $\tau=0.50$. For \emph{original} abstracts, 2013 to 2015 reports proxy FPR; 2023 to 2025 reports flag rate.}
 \label{tab:error_rates}
\Description{Rows compare Pangram, GPTZero, and an LLM-assisted baseline in two periods; columns report original false-positive or flag rates and rewrite false-negative rates before and after humanization. After humanization, false-negative rates reach 96.0--98.5 percent for the commercial detectors, compared with 19.7--30.5 percent for the baseline.}
\end{table}
Table~\ref{tab:error_rates} summarizes error rates at $\tau=0.50$. Under proxy labels, 2013 to 2015 originals incur FPR $=0\%$; in 2023 to 2025, original flag rates are 14.9\% (Pangram) and 8.9\% (GPTZero), not interpretable as confirmed false positives because published abstracts may already include LLM assistance. Non-STEM flag rates exceed STEM ($p<0.001$; Table~\ref{tab:domain_risk}). \emph{Refine (abstract only)} flag rates equal $1-\mathrm{FNR}$ in Table~\ref{tab:fnr_by_variant} (same scores, opposite proxy labels; Section~\ref{sec:error_operationalization}): Pangram 64.4\% (2013--15) and 80.1\% (2023--25); GPTZero 37.6\% and 48.5\%. After humanization, fewer than 4\% of AI-labeled rewrites remain detectable for AI detectors (FNR $>96\%$). The asymmetry is policy-critical: guideline-compliant editing triggers high flag rates, while humanizer evasion drops detection on synthetic text to near zero; honest editing currently faces more sanction risk than malicious evasion.

\begin{table}[t]
\centering
\small
\begin{tabular}{@{}lrrr@{}}
\toprule
Period & Pangram & GPTZero & $n$ \\
\midrule
2013--15 & 64.4\% & 37.6\% & 306 \\
2023--25 & 80.1\% & 48.5\% & 336 \\
\bottomrule
\end{tabular}
\caption{AI-assisted false-positive risk: pre-humanization flag rates on \emph{refine (abstract only)} outputs under proxy human-source labels at $\tau=0.50$. These are not FPR on unmodified originals.}
\label{tab:assisted_fp_risk}
\Description{Flag rates on abstract-only refinements rise from 64.4 to 80.1 percent for Pangram and from 37.6 to 48.5 percent for GPTZero between the earlier and later periods. These rates concern rewritten human-source texts, not false positives on unmodified originals.}
\end{table}

Table~\ref{tab:assisted_fp_risk} reports this proxy AI-assisted false-positive risk directly: 64--80\% for Pangram and 38--49\% for GPTZero. Table~\ref{tab:fnr_by_variant} decomposes pre-humanization FNR by rewrite condition. Lighter edits retain the highest FNR, especially on GPTZero; \emph{new (article only)} rewrites already exceed $\tau$ for most Pangram scores (2.6 to 2.7\% FNR). The LLM-assisted baseline flags 70.8\% of 2023 to 2025 originals and is treated as supplementary only.

\begin{table}[t]
\centering
\small
\setlength{\tabcolsep}{3pt}
\begin{tabular}{@{}llrrr@{}}
\toprule
Period & Condition & Pangram FNR & GPTZero FNR & $n$ \\
\midrule
2013-15 & refine (abs.\ only) & 35.6\% & 62.4\% & 306 \\
2013-15 & refine (abs.+paper) & 19.9\% & 54.6\% & 306 \\
2013-15 & new (article only) & 2.6\% & 15.7\% & 306 \\
2023-25 & refine (abs.\ only) & 19.9\% & 51.5\% & 336 \\
2023-25 & refine (abs.+paper) & 17.9\% & 48.8\% & 336 \\
2023-25 & new (article only) & 2.7\% & 14.9\% & 336 \\
\bottomrule
\end{tabular}
\caption{Pre-humanization FNR by rewrite condition at $\tau=0.50$.}
\label{tab:fnr_by_variant}
\Description{False-negative rates before humanization are listed by period, rewrite condition, and detector. Abstract-only refinement has the highest false-negative rates; new abstracts generated from articles have the lowest, at 2.6--2.7 percent for Pangram and 14.9--15.7 percent for GPTZero.}
\end{table}

\begin{table}[t]
\centering
\small
\setlength{\tabcolsep}{3pt}
\begin{tabular}{@{}lrrr@{}}
\toprule
Domain & Original flagged & Refine (abs.\ only) flagged & $n$ \\
\midrule
Chemistry & 2.1 / 1.1\% & 73.7 / 46.3\% & 95 \\
Computer science & 9.6 / 6.0\% & 69.9 / 36.1\% & 83 \\
Political science & 24.1 / 15.2\% & 86.1 / 54.4\% & 79 \\
Theology & 26.6 / 15.2\% & 92.4 / 58.2\% & 79 \\
\bottomrule
\end{tabular}
\caption{Domain-level flag rates in 2023 to 2025 (Pangram / GPTZero).}
\label{tab:domain_risk}
\Description{Each domain row gives original and abstract-only refinement flag rates, with Pangram followed by GPTZero, and the number of abstracts. Chemistry has the lowest original flag rates, while political science and theology have higher original and refinement flag rates than either STEM domain.}
\end{table}

\begin{figure}[t]
  \centering
  \includegraphics[width=\linewidth]{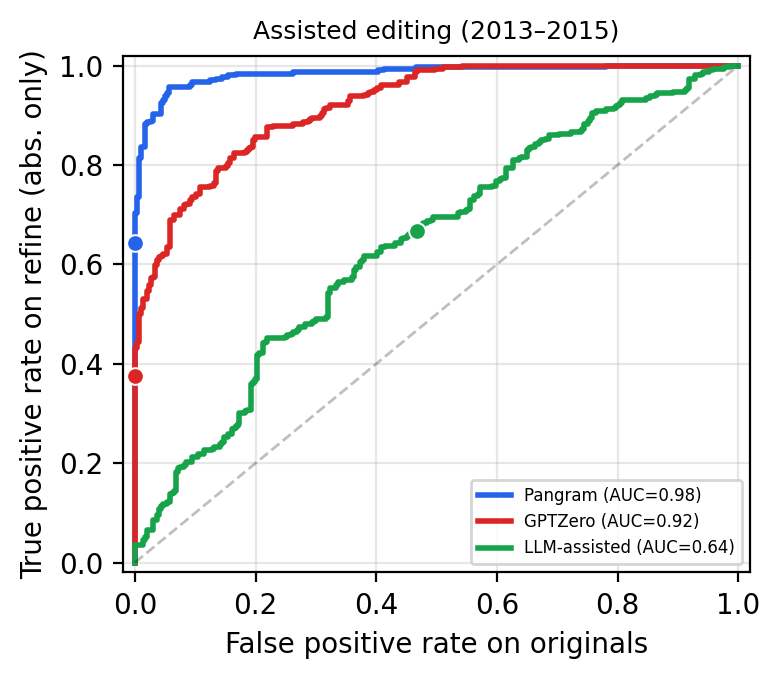}
  \caption{Assisted-editing ROC (2013-2015): \emph{original} vs.\ \emph{refine (abstract only)}; $\tau=0.50$}
  \label{fig:assisted_editing_roc}
\Description{ROC curves plot the true-positive rate on abstract-only refinements against the false-positive rate on 2013--2015 originals. Pangram, GPTZero, and the LLM-assisted baseline have areas under the curve of 0.98, 0.92, and 0.64; circles mark the threshold of 0.50, and a dashed diagonal denotes chance discrimination.}
\end{figure}

\begin{figure}[t]
  \centering
  \includegraphics[width=\linewidth]{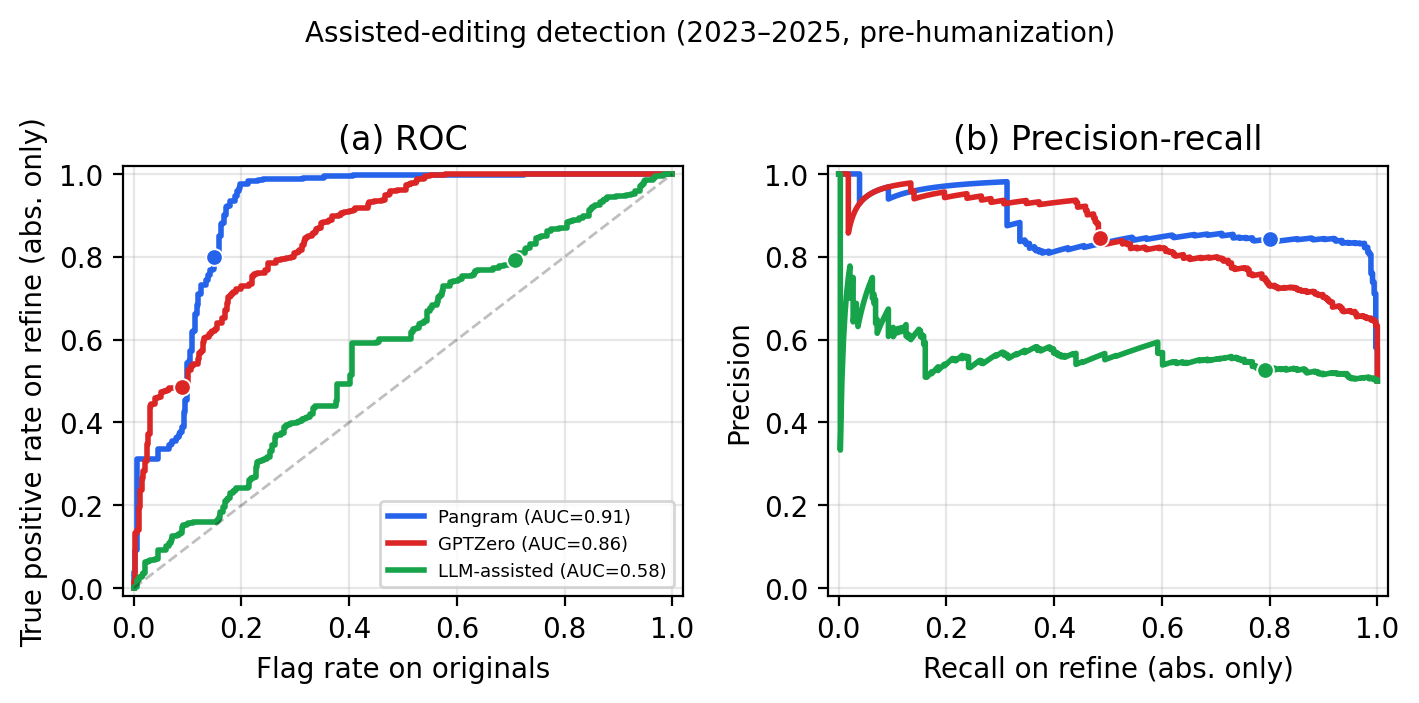}
  \caption{Assisted-editing ROC/PR (2023-2025, pre-humanization). Original-axis rates are flag rates, not proxy FPR.}
  \label{fig:assisted_editing_roc_pr_recent}
\Description{Two panels compare Pangram, GPTZero, and the LLM-assisted baseline on 2023--2025 abstract-only refinements: ROC curves on the left and precision-recall curves on the right. The ROC horizontal axis reports flags on originals with unknown AI-use histories; Pangram has the highest ROC area, and circles mark the threshold of 0.50.}
\end{figure}

Flag rates on 2023 to 2025 originals vary sharply by domain at the same $\tau$ (Table~\ref{tab:domain_risk}): Pangram flags 2.1\% in chemistry versus 26.6\% in theology; GPTZero shows the same ordering (1.1\% to 15.2\%). Vendor marketing claims sub-1\% false-positive rates on benchmark evaluations \cite{spero2025pangram_ai_assistance,gptzero2026benchmarking}; our corpus flags 9 to 15\% of recent \emph{original} abstracts and, on light edits, 80\% (Pangram) / 49\% (GPTZero) in 2023--25 and 64\% / 38\% in 2013--15 (flag rate $=1-\mathrm{FNR}$ in Table~\ref{tab:fnr_by_variant}; Table~\ref{tab:threshold_sensitivity}). Figure~\ref{fig:assisted_editing_roc} plots assisted-editing ROC on 2013 to 2015 abstracts (0\% original FPR at $\tau=0.50$). Pangram reaches AUC-ROC $0.98$ versus GPTZero $0.92$; at $\tau=0.50$ Pangram flags 64\% of light edits with no originals flagged, whereas GPTZero flags 38\%. Figure~\ref{fig:assisted_editing_roc_pr_recent} replicates the task on 2023 to 2025: original-axis rates are flag rates (15\% and 9\%), Pangram AUC-ROC is $0.91$, and 80\% of light edits are flagged at $\tau=0.50$.

\begin{figure}[t]
    \centering
    \includegraphics[width=\linewidth]{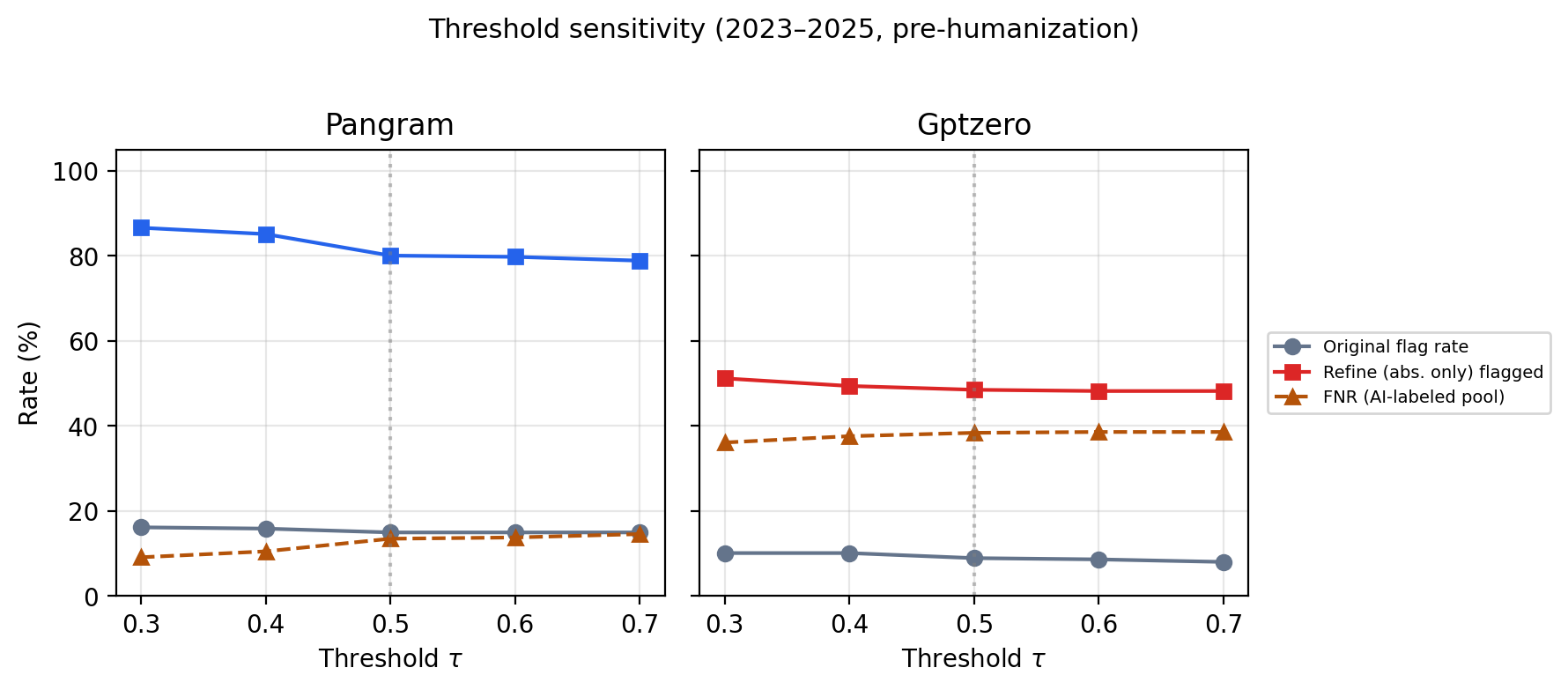}
    \caption{Threshold trade-off (2023 to 2025, pre-humanization). Three policy rates vs.\ $\tau$ for Pangram (left) and GPTZero (right). Dotted vertical line: $\tau=0.50$.}
    \label{fig:threshold_tradeoff}
\Description{Pangram and GPTZero panels plot percentage rates against thresholds from 0.3 to 0.7. Circles denote original flag rates, squares denote abstract-only refinement flag rates, and dashed lines with triangles denote pooled rewrite false-negative rates; increasing the threshold lowers flags and raises misses, with dotted vertical lines marking 0.50.}
\end{figure}

Figure~\ref{fig:threshold_tradeoff} extends the threshold sweep to 0.3--0.7 for both detectors. It shows original and assisted-writing flag rates alongside FNR on all three LLM rewrite conditions; flags on recent originals are not confirmed false positives. No threshold $\tau \in \{0.4,0.5,0.6\}$ simultaneously keeps original flag rates, refine capture, and AI-pool FNR low (Table~\ref{tab:threshold_sensitivity}): Pangram holds $\approx$15\% original flags and $\approx$80\% refine flags across that band, while raising $\tau$ from 0.4 to 0.6 slightly \emph{increases} FNR on AI-labeled rewrites (10.5\% to 13.8\%).

\begin{table}[t]
\centering
\small
\setlength{\tabcolsep}{3pt}
\begin{tabular}{@{}llrrr@{}}
\toprule
Detector & $\tau$ &
\makecell{Flag rate\\(original)} &
\makecell{FNR\\(AI-labeled)} &
\makecell{Refine\\flagged} \\
\midrule
Pangram & 0.4 & 15.8\% & 10.5\% & 85.1\% \\
Pangram & 0.5 & 14.9\% & 13.5\% & 80.1\% \\
Pangram & 0.6 & 14.9\% & 13.8\% & 79.8\% \\
GPTZero & 0.4 & 10.1\% & 37.6\% & 49.4\% \\
GPTZero & 0.5 & 8.9\% & 38.4\% & 48.5\% \\
GPTZero & 0.6 & 8.6\% & 38.6\% & 48.2\% \\
\bottomrule
\end{tabular}
\caption{Threshold sensitivity (2023-2025, pre-humanization).}
\label{tab:threshold_sensitivity}
\Description{Rows compare thresholds of 0.4, 0.5, and 0.6 for Pangram and GPTZero; columns give original flag rates, pooled rewrite false-negative rates, and abstract-only refinement flag rates. Higher thresholds reduce flag rates and increase false-negative rates, with little change between 0.5 and 0.6.}
\end{table}

\subsection{Linguistic Correlates}
Long-token and AWL ratios correlate positively with 2023 to 2025 scores ($\rho \approx 0.30$ to $0.35$; $p<0.001$); numeric and non-alphabetic ratios correlate negatively (Table~\ref{tab:feature_corr_summary}). After centering within domain, long-token and AWL associations strengthen for Pangram ($r>0.41$). After adjustment for length, AWL correlations remain positive (Pangram/GPTZero: $0.337/0.261$), as do long-token correlations ($0.321/0.276$). Length alone therefore does not account for these associations. These analyses pool originals and rewrites; they do not isolate associations within unmodified human writing or establish causal effects.

\begin{table}[t]
\centering
\small
\setlength{\tabcolsep}{4pt}
\begin{tabular}{@{}lrr@{}}
\toprule
Feature & Pangram $\rho$ & GPTZero $\rho$ \\
\midrule
AWL token ratio & 0.346 & 0.320 \\
Long token ratio & 0.336 & 0.304 \\
Non-alphabetic token ratio & $-0.149$ & $-0.197$ \\
Numeric token ratio & $-0.146$ & $-0.171$ \\
\bottomrule
\end{tabular}
\caption{Strongest score-feature associations (2023 to 2025, pre-humanization; Spearman $\rho$, both detectors $p<0.001$).}
\label{tab:feature_corr_summary}
\Description{Spearman correlations between four text features and detector scores are reported for Pangram and GPTZero. Academic Word List density and long-token ratio have positive correlations of 0.304--0.346, while numeric and non-alphabetic token ratios have negative correlations of -0.146 to -0.197.}
\end{table}

\subsection{Humanization and Detector Agreement}
Post-humanization Pangram means fall to 0.03 to 0.04 across conditions in 2023 to 2025, alongside the high FNR in Table~\ref{tab:error_rates}. Pangram and GPTZero agree moderately before humanization but weakly afterward (Table~\ref{tab:detector_agreement}). Across $n=2{,}568$ humanized pairs, long-token ratio, AWL density, and lexical diversity fall ($p<0.001$; Table~\ref{tab:humanization_linguistics}). These changes are consistent with reduced use of features associated with higher scores, but do not establish how the proprietary humanizer achieves evasion.

\begin{table}[t]
\centering
\small
\setlength{\tabcolsep}{4pt}
\begin{tabular}{@{}llrr@{}}
\toprule
Period & Phase & Spearman $\rho$ & Cohen's $\kappa$ \\
\midrule
2013-15 & pre & 0.721 & 0.500 \\
2013-15 & post & 0.100 & 0.200 \\
2023-25 & pre & 0.736 & 0.539 \\
2023-25 & post & 0.188 & 0.221 \\
\bottomrule
\end{tabular}
\caption{Pangram and GPTZero agreement ($n=1{,}224$ pairs in 2013--15; $n=1{,}344$ in 2023--25).}
\label{tab:detector_agreement}
\Description{Spearman score correlations and Cohen's kappa compare Pangram with GPTZero before and after humanization in both periods. Spearman correlations fall from 0.721--0.736 to 0.100--0.188, and kappa falls from 0.500--0.539 to 0.200--0.221.}
\end{table}

\begin{table}[t]
\centering
\small
\setlength{\tabcolsep}{3pt}
\begin{tabular}{@{}lrrrr@{}}
\toprule
Feature & Pre & Post & $\Delta$ & $p$ \\
\midrule
Long-token ratio & 0.184 & 0.138 & $-0.045$ & $<0.001$ \\
AWL token ratio & 0.157 & 0.121 & $-0.036$ & $<0.001$ \\
Type-token ratio & 0.700 & 0.605 & $-0.094$ & $<0.001$ \\
Word count & 180 & 229 & $+49$ & $<0.001$ \\
\bottomrule
\end{tabular}
\caption{Pooled linguistic shifts after Undetectable AI v11 humanization ($n=2{,}568$ pairs).}
\label{tab:humanization_linguistics}
\Description{Mean values before and after humanization, paired changes, and permutation-test p-values are listed for four features across 2,568 pairs. Long-token ratio, Academic Word List density, and type-token ratio decrease, while mean word count rises from 180 to 229; all reported p-values are below 0.001.}
\end{table}

\section{Discussion}

Before humanization, detectors can distinguish LLM-modified abstracts from originals, especially Pangram (AUC-ROC $0.98$ in 2013--2015 and $0.91$ in 2023--2025). This sensitivity concerns traces of model intervention in the wording, not ownership of the ideas. Both abstract-only refinement and generation from an article frequently trigger flags, despite different model inputs and writing tasks. These scores alone cannot reconstruct the edits or quantify how much of the ideas, original analysis, or effort belongs to the author rather than the model; our experiment does not measure those contributions. If used at all, detectors should be one input to a broader review of drafts, revision histories, disclosed AI use, and the author's explanation of the work.

Detector errors are structured, not random: long-token and AWL density track higher scores in non-STEM prose, while symbol-heavy STEM text scores lower even on AI-assisted rewrites. Under proxy labels, compliant editing triggers flags while humanization drives detection on AI-labeled text below 4\%. Institutions therefore face a catch-22: prohibit all AI-assisted editing and deny legitimate support to multilingual and novice writers, or enforce with tools that punish acceptable use and miss evasion at scale \cite{wallwork2025ai,sun2026trusting}. Uniform detector policies may also raise differential-impact concerns when flag rates differ by discipline \cite{zdravkova2024preventing}.

\paragraph{Practice} These patterns create concrete educational risk: students may be flagged while drafting their own ideas and using AI only for clarity or grammar, yet others who generate inorganic drafts can evade detection after humanization. Multilingual learners face additional harm when AI-assisted translation or editing elevates scores on the student's own work \cite{zhang2023don}. Detector output must be treated as weak evidence, not default misconduct evidence. Any flag must be paired with drafting history and sanctions shall be reserved for cases where human judgment supports a lack of appropriate student effort. Expect higher rates in non-STEM prose. Finally, FERPA/GDPR guidelines must be followed before uploading students' confidential work to third-party services \cite{quinnipiac_ai_detectors}.

\paragraph{Limitations} We analyze 642 English published abstracts, not student essays, and do not evaluate translation, groups of multilingual writers, or actual sanctions. Labels are proxies; recent originals have unknown AI-use histories, and refine outputs are not verified grammar-only edits. We used one rewrite generator (Gemini 3 Flash), two commercial detectors (Pangram 3.2, GPTZero), one humanizer configuration (Undetectable AI v11), and one LLM-assisted baseline. Results may differ across genres, languages, models, and service versions. Feature associations are correlational. We did not assess whether rewriting or humanization preserved scientific meaning; low detector scores alone do not establish output quality.

\section{Conclusion}
Commercial detectors on published abstracts punish acceptable AI-assisted editing, compound risk for English-language learners who use AI for translation, and fail to catch humanized synthetic text. Light edits are flagged at 64--80\% by Pangram and 38--49\% by GPTZero. After humanization, more than 96\% of AI-labeled rewrites evade Pangram and GPTZero. \textbf{Together, these results define an \emph{integrity catch-22}: high detector flags on assisted writing alongside near-total misses after humanization.} Integrity programs therefore need transparent AI-use norms and process evidence, not standalone detector scores.

\begin{acks}
We would like to thank Pangram for the use of their API, as well as Notre Dame Teaching \& Learning, and the Notre Dame AI Council.
\end{acks}

\printbibliography

\ifincludeappendices
\appendix

\section*{Appendix}

\section{Limitations}
\label{app:limitations}

\textbf{Language and corpus setting:}
Our dataset is limited to English. We do not evaluate AI detection or humanization for other languages, so our findings should not be extrapolated to multilingual classrooms or non-English scholarly writing without further study. We analyze published research abstracts drawn from four OpenAlex domains (chemistry, computer science, political science, and theology), not student essays or other classroom genres. Detector behavior on shorter, pedagogically shaped student writing may differ from the abstract-length prose studied here.

\textbf{Labels and measurement:}
Ground truth is experimental rather than adjudicated. We treat unmodified \emph{original} abstracts as proxy human writing and LLM rewrites as proxy AI writing. For 2023 to 2025 originals, real-world AI assistance is unobserved, so detector positives on those texts are flag rates, not confirmed false positives. All reported false-positive and false-negative rates depend on this labeling scheme and on a fixed score threshold ($\tau=0.50$). Text-feature associations are correlational and do not establish causal mechanisms.

\textbf{Models, detectors, and humanization:}
We used one rewrite generator (Gemini 3 Flash), two commercial API detectors (Pangram 3.2 and GPTZero), one humanizer configuration (Undetectable AI v11), and one supplementary LLM-assisted baseline (GPT-5 Nano). Results may not generalize to other models, detector versions, humanizers, or prompting strategies. Post-humanization evasion claims in the main text concern Pangram and GPTZero. The LLM-assisted baseline was also scored after humanization (Table~\ref{tab:error_rates}) but remains supplementary: it still flags most humanized rewrites (post-humanization FNR 30.5\% and 19.7\%).

\textbf{Corpus size and cost:}
We targeted 800 abstracts (100 per domain-time bucket) and retained 642 after length and PDF filtering. Corpus size is constrained by the cost of large-scale API detection and paid humanization. Each retained paper requires multiple rewrite calls, pre-detection passes across three scorers, one humanization credit per variant, and post-detection with the same three scorers: Pangram, GPTZero, and the LLM-assisted baseline. A multi-model rewrite sweep would multiply those costs without changing the core error-analysis design.

\textbf{Future work:}
Future studies could expand to non-English corpora and to adjudicated student writing with verified authorship and AI-use histories. Developing local or open models for rewriting, detection, and humanization could reduce reliance on paid APIs and enable larger, more diverse datasets. Such systems would need careful validation before use in high-stakes integrity workflows, but they offer a path toward reproducible benchmarks at scales that are impractical under current commercial pricing. An open question is whether prose shaped by frontier models (or by newer assistance tools) systematically elevates commercial detector scores and therefore false-positive risk on human writing relative to earlier LLM generations; our 2023 to 2025 window cannot separate model vintage from unobserved real-world AI use on \emph{original} abstracts, but targeted comparisons across model families on controlled human-authored drafts would clarify that risk.

\section{Dual interpretation of light editing}
\label{app:dual_labels}

Reviewers and integrity offices need a clear account of why \emph{refine (abstract only)} appears in two rate families.

For each source paper we generate a Gemini~3 Flash rewrite that edits only the abstract. Detector score $s$ and flag $\hat{y}=\mathbb{I}\{s\ge\tau\}$ are identical regardless of interpretation.

\paragraph{Reading A: proxy-AI labeling.}
Label $y=\mathrm{AI}$. Then $\mathrm{FN}=\mathbb{I}\{\hat{y}=0\}$ and FNR is the miss rate among AI-labeled rewrites. This reading supports detection-evaluation claims (can the tool catch LLM text?).

\paragraph{Reading B: human-source / assistance.}
Label the intellectual source as human and treat the LLM as a polish tool. Then the quantity of interest is the \emph{assisted-writing flag rate} $P(\hat{y}=1\mid \text{refine abs.\ only})$. This reading supports policy claims about sanction risk for authors who retain ownership of claims. It is \textbf{not} proxy FPR on unmodified human text, and it must not be abbreviated as FPR in tables or prose.

\paragraph{Why the rates are not comparable as ordinary errors:}
FNR under Reading~A and assisted-writing flag rate under Reading~B use opposite label assignments for the same texts. Comparing Pangram's 64--80\% light-edit flag rates to $>96\%$ false negatives as if they were cells of one confusion matrix would be a category error. We therefore report both quantities, name them differently, and interpret the integrity catch-22 as a \emph{policy} asymmetry across readings, not as a single-threshold accuracy summary.

\paragraph{Relationship to institutional rules:}
Policies that ban all generative AI would treat Reading~A as decisive. Policies that allow grammar/clarity assistance but ban ghostwriting sit between the readings and require process evidence our detectors do not observe. Our refine prompt is closer to light generative rewriting than to grammar-only editing; assisted-writing flag rates should be read accordingly.

\section{Paper-clustered inference}
\label{app:clustered_inference}

Each source paper contributes an original abstract and three rewrite conditions. Permutation tests and bootstrap intervals that resample instances can understate uncertainty when within-paper dependence is ignored. Therefore, we report analyses that resample at the \texttt{paper\_id} level (2{,}000 bootstrap replicates or 5{,}000 permutations; seed 42; $\tau=0.50$).

\textbf{Error and flag rates:}
Table~\ref{tab:clustered_error_cis} shows paper-clustered 95\% intervals. Point estimates match the main text (e.g., 2023 to 2025 pre-humanization Pangram original flag rate 14.9\% $[11.0, 18.8]$; refine assisted-writing flag rate 80.1\% $[75.9, 84.5]$; AI-pool FNR 13.5\% $[11.0, 16.0]$; post-humanization AI-pool FNR 96.6\% $[95.2, 97.8]$). Clustering widens intervals relative to naive instance bootstrap but does not change substantive conclusions.

\textbf{STEM vs.\ non-STEM:}
Table~\ref{tab:clustered_stem} uses one original-abstract score per paper. In 2023 to 2025, non-STEM flag rates remain far above STEM (Pangram 25.3\% vs.\ 5.6\%; GPTZero 15.2\% vs.\ 3.4\%; paper-label permutation $p=0.0002$ for both detectors).

\textbf{Feature associations:}
Table~\ref{tab:clustered_feature_corr} retains instance-level Spearman $\rho$ (matching Table~\ref{tab:feature_corr_summary}) but forms CIs by resampling papers. Long-token and AWL associations remain positive with intervals excluding zero for both detectors (e.g., Pangram AWL $\rho=0.346$ $[0.298, 0.395]$). Paper-averaged $\rho$ is attenuated because averaging mixes originals and rewrites within paper; we report it for transparency, not as a replacement for the instance-level estimand.

\begin{table*}[h]
\centering
\small
\begin{tabular}{lllrrrr}
\toprule
Period & Phase & Detector & Metric & Point & 95\% CI (paper) & $n_{\mathrm{papers}}$ \\
\midrule
2013-2015 & post & gptzero & FNR (AI-labeled pool) & 96.0\% & [94.6\%, 97.3\%] & 306 \\
2013-2015 & post & gptzero & Original flag rate & 1.3\% & [0.3\%, 2.6\%] & 306 \\
2013-2015 & post & gptzero & Refine (abs.\ only) flag rate & 3.3\% & [1.3\%, 5.6\%] & 306 \\
2013-2015 & post & pangram & FNR (AI-labeled pool) & 98.5\% & [97.5\%, 99.2\%] & 306 \\
2013-2015 & post & pangram & Original flag rate & 0.3\% & [0.0\%, 1.0\%] & 306 \\
2013-2015 & post & pangram & Refine (abs.\ only) flag rate & 3.3\% & [1.6\%, 5.6\%] & 306 \\
2013-2015 & pre & gptzero & FNR (AI-labeled pool) & 44.2\% & [40.6\%, 47.8\%] & 306 \\
2013-2015 & pre & gptzero & Original flag rate & 0.0\% & [0.0\%, 0.0\%] & 306 \\
2013-2015 & pre & gptzero & Refine (abs.\ only) flag rate & 37.6\% & [32.0\%, 43.1\%] & 306 \\
2013-2015 & pre & pangram & FNR (AI-labeled pool) & 19.4\% & [16.4\%, 22.3\%] & 306 \\
2013-2015 & pre & pangram & Original flag rate & 0.0\% & [0.0\%, 0.0\%] & 306 \\
2013-2015 & pre & pangram & Refine (abs.\ only) flag rate & 64.4\% & [58.8\%, 69.3\%] & 306 \\
2023-2025 & post & gptzero & FNR (AI-labeled pool) & 96.1\% & [94.6\%, 97.3\%] & 336 \\
2023-2025 & post & gptzero & Original flag rate & 1.5\% & [0.3\%, 3.0\%] & 336 \\
2023-2025 & post & gptzero & Refine (abs.\ only) flag rate & 3.9\% & [1.8\%, 6.0\%] & 336 \\
2023-2025 & post & pangram & FNR (AI-labeled pool) & 96.6\% & [95.2\%, 97.8\%] & 336 \\
2023-2025 & post & pangram & Original flag rate & 3.3\% & [1.5\%, 5.4\%] & 336 \\
2023-2025 & post & pangram & Refine (abs.\ only) flag rate & 3.6\% & [1.8\%, 5.7\%] & 336 \\
2023-2025 & pre & gptzero & FNR (AI-labeled pool) & 38.4\% & [34.7\%, 41.9\%] & 336 \\
2023-2025 & pre & gptzero & Original flag rate & 8.9\% & [6.2\%, 11.9\%] & 336 \\
2023-2025 & pre & gptzero & Refine (abs.\ only) flag rate & 48.5\% & [43.2\%, 53.6\%] & 336 \\
2023-2025 & pre & pangram & FNR (AI-labeled pool) & 13.5\% & [11.0\%, 16.0\%] & 336 \\
2023-2025 & pre & pangram & Original flag rate & 14.9\% & [11.0\%, 18.8\%] & 336 \\
2023-2025 & pre & pangram & Refine (abs.\ only) flag rate & 80.1\% & [75.9\%, 84.5\%] & 336 \\
\bottomrule
\end{tabular}
\caption{Paper-clustered bootstrap intervals ($n_{\mathrm{boot}}=2{,}000$; seed 42) at $\tau=0.50$.}
\label{tab:clustered_error_cis}
\end{table*}

\begin{table*}[h]
\centering
\small
\begin{tabular}{llrrrrr}
\toprule
Period & Detector & $\Delta$ mean & Flag STEM & Flag non-STEM & $n$ STEM/non & $p$ (paper) \\
\midrule
2013-2015 & gptzero & $-0.006$ & 0.0\% & 0.0\% & 154/152 & 0.026 \\
2013-2015 & pangram & $0.001$ & 0.0\% & 0.0\% & 154/152 & 0.60 \\
2023-2025 & gptzero & $0.110$ & 3.4\% & 15.2\% & 178/158 & 0.0002 \\
2023-2025 & pangram & $0.202$ & 5.6\% & 25.3\% & 178/158 & 0.0002 \\
\bottomrule
\end{tabular}
\caption{STEM vs.\ non-STEM on \emph{original} abstracts.}
\label{tab:clustered_stem}
\end{table*}

\begin{table*}[h]
\centering
\small
\begin{tabular}{llrrrr}
\toprule
Detector & Feature & $\rho$ (inst.) & 95\% CI (paper) & $\rho$ (paper avg.) & $n_{\mathrm{papers}}$ \\
\midrule
gptzero & awl token ratio & 0.320 & [0.269, 0.370] & 0.139 & 336 \\
gptzero & long token ratio & 0.304 & [0.249, 0.359] & 0.079 & 336 \\
pangram & awl token ratio & 0.346 & [0.298, 0.395] & 0.148 & 336 \\
pangram & long token ratio & 0.336 & [0.290, 0.386] & 0.072 & 336 \\
\bottomrule
\end{tabular}
\caption{Text-feature associations (2023 to 2025, pre-humanization).}
\label{tab:clustered_feature_corr}
\end{table*}

\section{Abstract Rewriting Prompts}
\label{app:prompts}

We provide the prompts used to instruct the model for abstract rewriting.

\textbf{Prompt used for the Refine (abstract only) setting}:  You are an expert scientist who excels in writing papers. Your task is to rewrite a given paper abstract to sound better and more human-like. The abstract will be given in the next message. Please only return the rewritten text, do not return any instructions or anything like that. Only return the rewritten text. Make sure to make at least some edits or changes to the text.

\textbf{Prompt used for the Refine (abstract + article) setting}: You are an expert scientist who excels in writing papers. Your task is to rewrite a given paper abstract to sound better and more human-like, using the full text of the paper to inform your rewriting. The abstract and text will be given in the next message. Please only return the rewritten text, do not return any instructions or anything like that. Only return the rewritten text. Make sure to make at least some edits or changes to the text.

\textbf{Prompt used for the New (article only) setting}:  You are an expert scientist who excels in writing papers. Your task is to write a naturally sounding abstract for a paper that will be given to you. The full will be given in the next message. Please only return the written abstract, do not return any instructions or anything like that. Only return the written text. Make sure to not copy the text of the paper verbatim.

\section{Vendor marketing claims}
\label{app:vendor_claims}

We record the commercial claims that motivate academic-integrity adoption of AI detectors and humanizers. Table~\ref{tab:vendor_claims} organizes them in the same order as our framing: headline detector accuracy, claims about \emph{AI-assisted} versus fully generated writing, and humanizer evasion guarantees. All figures come from vendor marketing or vendor-run benchmarks (accessed December 2025 to May 2026), not from our corpus.

\subsection{Headline detector accuracy}
Vendors advertise near-perfect detection on benchmark-style evaluations. Pangram's Pangram 3.0 announcement reports 99.98\% accuracy on AI-generated text with near-zero false positives on the fully AI-generated label \cite{spero2025pangram_ai_assistance}. GPTZero's peer-reviewed detector paper reports 99.39\% accuracy on its multi-domain evaluation (Table 2 aggregate for GPTZero 4.1b) \cite{adam2026gptzero}. GPTZero's public benchmarking post additionally claims 99.76\% average accuracy with 0.08\% false-positive rate across four commercial domains \cite{gptzero2026benchmarking}.

\subsection{AI-assisted versus fully generated writing}
A separate set of claims concerns whether detectors can separate light editing from full synthesis. Pangram 3.0 markets explicit classes (fully human, lightly AI-assisted, moderately AI-assisted, fully AI-generated) \cite{spero2025pangram_ai_assistance}. GPTZero describes multiclass outputs that separate human text, mixed or lightly edited text, and pure AI \cite{gptzero2026benchmarking}. Independent work notes that accuracy-bias trade-offs and binarization choices can obscure assisted-writing risk \cite{pratama2025accuracy}.

\subsection{Humanizer evasion guarantees}
Detector adoption has spurred humanizers that target surface metrics such as perplexity and burstiness \cite{wallwork2025ai}. Undetectable AI's pricing page states a money-back guarantee: if humanized output is flagged as not human, the humanization cost is refunded \cite{undetectable_ai_pricing}. Its marketing site also advertises 99\%+ detector accuracy, a Forbes ``\#1 Best AI Detector'' rating (2024), and more than 22 million users as of March 2026 \cite{undetectable_ai_home}.

\begin{table*}[h]
\centering

\small
\begin{tabular}{p{0.17\linewidth}p{0.20\linewidth}p{0.43\linewidth}p{0.12\linewidth}}
\toprule
Vendor & Claim type & Stated claim & Source \\
\midrule
Pangram & Accuracy & 99.98\% accuracy on AI-generated text; near-zero false positives on the AI-generated label. & \cite{spero2025pangram_ai_assistance} \\
GPTZero & Accuracy & 99.39\% accuracy on multi-domain detector evaluation (GPTZero 4.1b). & \cite{adam2026gptzero} \\
GPTZero & Accuracy (marketing) & 99.76\% average accuracy; 0.08\% FPR; 99.60\% recall (vendor benchmark, v4.3b). & \cite{gptzero2026benchmarking} \\
Pangram & Assisted vs.\ generated & Four-way taxonomy: human, light assistance, moderate assistance, fully AI-generated. & \cite{spero2025pangram_ai_assistance} \\
GPTZero & Assisted vs.\ generated & Multiclass taxonomy (human, mixed/edited, pure AI) for benchmarking. & \cite{gptzero2026benchmarking} \\
Literature & Assisted vs.\ generated & Detectors face accuracy-bias trade-offs; binarized outputs can hide assisted-writing risk. & \cite{pratama2025accuracy} \\
Undetectable AI & Humanizer guarantee & Money-back if humanized text is ``flagged as not human.'' & \cite{undetectable_ai_pricing} \\
Undetectable AI & Marketing & 99\%+ detector accuracy; Forbes \#1 (2024); 22M+ users (March 2026). & \cite{undetectable_ai_home} \\
\bottomrule
\end{tabular}
\caption{Vendor claims cited in our framing (marketing and vendor benchmarks).}
\label{tab:vendor_claims}
\end{table*}

\textbf{Contrast with our corpus:}
These claims are not reproduced on published English abstracts under our proxy labels at $\tau=0.50$. Pangram and GPTZero flag 14.9\% and 8.9\% of 2023 to 2025 \emph{original} abstracts (Table~\ref{tab:error_rates}), inconsistent with sub-1\% false-positive marketing. \emph{Refine (abstract only)} outputs (human-source, light LLM edit) are flagged 64 to 80\% of the time by Pangram and 38 to 49\% by GPTZero; Table~\ref{tab:fnr_by_variant}, showing that assisted-writing risk is substantial even when vendors market assistance-level detection. After Undetectable AI v11 humanization, false-negative rates on AI-labeled rewrites exceed 96\% for both Pangram and GPTZero, so humanized text routinely bypasses detectors despite the refund guarantee. Domain-level dispersion (Tables~\ref{tab:pangram_domain_pre_2025} and~\ref{tab:pangram_domain_post_2025}) further shows that pooled benchmark accuracy can mask large field-to-field variance.

\section{Definitions of text ratios}
\label{app:text_ratio_definitions}

We compute several token-level ratios from each abstract to probe whether surface properties correlate with detector scores. Let $t$ be the abstract text (a string). We extract an ordered token sequence $T(t)=(\tau_1,\ldots,\tau_n)$ using the regular expression \begin{verbatim} [A-Za-z0-9][A-Za-z0-9\-\+\./]* \end{verbatim}, and let $n=|T(t)|$. For convenience, we use an indicator function $\mathbb{I}\{\cdot\}$ that equals 1 when its condition holds and 0 otherwise. If $n=0$, the ratios below are undefined.

\paragraph{Numeric token ratio.}
\[
\mathrm{numeric\_token\_ratio}(t) = \frac{1}{n}\sum_{i=1}^{n} \mathbb{I}\{\exists c \in \tau_i:\; c \in \{0,\ldots,9\}\}.
\]

\paragraph{Non-alphabetic token ratio.}
\[
\mathrm{nonalpha\_token\_ratio}(t) = \frac{1}{n}\sum_{i=1}^{n} \mathbb{I}\{\exists c \in \tau_i:\; c \text{ is not a letter}\}.
\]
This captures tokens that include punctuation such as hyphens, periods, slashes, or plus signs (as well as digits).

\paragraph{Long token ratio.}
\[
\mathrm{long\_token\_ratio}(t) = \frac{1}{n}\sum_{i=1}^{n} \mathbb{I}\{|\tau_i| \ge 10\}.
\]

\paragraph{Acronym ratio.}
\[
\mathrm{acronym\_ratio}(t) = \frac{1}{n}\sum_{i=1}^{n} \mathbb{I}\{\tau_i \text{ is uppercase and } |\tau_i|\ge 2\}.
\]

\paragraph{AWL token ratio.}
Let $A$ be Coxhead's Academic Word List (AWL) headwords \cite{coxhead2000new} (570 items). We match tokens by Porter stemming against this list: for each token $\tau$, we form a letters-only normalization $\mathrm{alpha}(\tau)$ by removing non-letters, then compute its Porter stem. Let $S(A)$ be the set of Porter stems of the AWL headwords (plus a small set of US spelling variants for headwords whose Porter stems differ, e.g., \textit{analyse}/\textit{analyze}).
\[
\mathrm{awl\_token\_ratio}(t) = \frac{1}{n}\sum_{i=1}^{n} \mathbb{I}\{\mathrm{stem}(\mathrm{alpha}(\tau_i)) \in S(A)\}.
\]

\paragraph{Type-token ratio (lexical diversity).}
Let $V(t) = \{ \mathrm{lower}(\tau) : \tau \in T(t) \}$ be the set of unique case-folded tokens. Then
\[
\mathrm{ttr}(t) = \frac{|V(t)|}{n}.
\]

\paragraph{Word count.}
We report word counts using whitespace splitting: $\mathrm{words}(t)=|\mathrm{split}(t)|$.

\paragraph{Average words per sentence.}
We split sentences by punctuation (\texttt{.}, \texttt{!}, \texttt{?}) and compute the average number of whitespace-delimited words per resulting sentence span.

\paragraph{Short-sentence ratio.}
Using the same sentence splits, let $m$ be the number of sentences and $w_j$ the word count of sentence $j$. Then
\[
\mathrm{short\_sentence\_ratio}(t) = \frac{1}{m}\sum_{j=1}^{m} \mathbb{I}\{w_j \le 8\}.
\]
We use eight words as a simple fragmentation proxy in the humanization mechanism analysis.

\FloatBarrier
\section{Supplementary numerical results}
\label{app:results_tables}

The tables below support the main-text results by research question. Tables~\ref{tab:pangram_ranges} through~\ref{tab:pangram_domain_post_2025} provide pooled and domain-level Pangram summaries (pre- and post-humanization). For corpus counts see Table~\ref{tab:corpus_counts} in the main text.

\subsection{Pangram score summaries}
Tables~\ref{tab:pangram_ranges} and~\ref{tab:pangram_ranges_post} pool all domains; Tables~\ref{tab:pangram_domain_pre_2015} through~\ref{tab:pangram_domain_post_2025} decompose the same statistics by domain. SD captures score dispersion within each cell; \emph{Outliers} counts scores outside $1.5\times\mathrm{IQR}$ fences (same rule as boxplot fliers). GPTZero and the LLM-assisted baseline follow similar orderings.
\begin{table*}[h]
\centering
\small
\begin{tabular}{lllrrrr}
\toprule
Period & Condition & Mean & SD & [P25, P75] & $n$ & Outliers \\
\midrule
2013 to 2015 & original & 0.002 & 0.022 & [0.000, 0.000] & 306 & 44 \\
2013 to 2015 & refine (abs.\ only) & 0.669 & 0.423 & [0.143, 1.000] & 306 & 0 \\
2013 to 2015 & refine (abs.+paper) & 0.817 & 0.351 & [0.927, 1.000] & 306 & 68 \\
2013 to 2015 & new (article only) & 0.975 & 0.136 & [1.000, 1.000] & 306 & 24 \\
2023 to 2025 & original & 0.155 & 0.352 & [0.000, 0.001] & 336 & 74 \\
2023 to 2025 & refine (abs.\ only) & 0.827 & 0.328 & [0.937, 1.000] & 336 & 77 \\
2023 to 2025 & refine (abs.+paper) & 0.840 & 0.325 & [0.995, 1.000] & 336 & 82 \\
2023 to 2025 & new (article only) & 0.974 & 0.142 & [1.000, 1.000] & 336 & 29 \\
\bottomrule
\end{tabular}
\caption{Pangram score summaries before humanization, pooled across domains (mean, SD, [IQR]; outliers outside $1.5\times\mathrm{IQR}$).}
\label{tab:pangram_ranges}
\end{table*}

\textbf{Domain variance before humanization.}
Pre-humanization dispersion is domain-dependent and explains the STEM/non-STEM contrasts in the main text. On \emph{original} abstracts in 2023 to 2025, political science and theology show the highest means and SD (Table~\ref{tab:pangram_domain_pre_2025}), matching elevated flag rates in Table~\ref{tab:domain_risk}; chemistry stays low on both metrics. For LLM rewrites, outliers concentrate in high-scoring conditions (\emph{refine (abstract + article)}, \emph{new (article only)}) where the IQR sits near 1.0, while \emph{refine (abstract only)} often has SD comparable across domains but wider score spread in computer science (lower quartile) than theology (tight upper quartile).

\begin{table*}[h]
\centering
\scriptsize
\begin{tabular}{lllrrrr}
\toprule
Domain & Condition & Mean & SD & [P25, P75] & $n$ & Outliers \\
\midrule
Chemistry & original & 0.003 & 0.020 & [0.000, 0.000] & 84 & 12 \\
 & refine (abs.\ only) & 0.698 & 0.413 & [0.250, 1.000] & 84 & 0 \\
 & refine (abs.+paper) & 0.810 & 0.358 & [0.921, 1.000] & 84 & 19 \\
 & new (article only) & 0.953 & 0.195 & [1.000, 1.000] & 84 & 9 \\
\addlinespace
Computer science & original & 0.000 & 0.000 & [0.000, 0.000] & 70 & 9 \\
 & refine (abs.\ only) & 0.567 & 0.438 & [0.040, 1.000] & 70 & 0 \\
 & refine (abs.+paper) & 0.687 & 0.442 & [0.074, 1.000] & 70 & 0 \\
 & new (article only) & 0.954 & 0.177 & [1.000, 1.000] & 70 & 9 \\
\addlinespace
Political science & original & 0.005 & 0.039 & [0.000, 0.000] & 78 & 14 \\
 & refine (abs.\ only) & 0.717 & 0.406 & [0.305, 1.000] & 78 & 0 \\
 & refine (abs.+paper) & 0.923 & 0.217 & [1.000, 1.000] & 78 & 18 \\
 & new (article only) & 0.994 & 0.052 & [1.000, 1.000] & 78 & 4 \\
\addlinespace
Theology & original & 0.000 & 0.000 & [0.000, 0.000] & 74 & 11 \\
 & refine (abs.\ only) & 0.682 & 0.431 & [0.124, 1.000] & 74 & 0 \\
 & refine (abs.+paper) & 0.835 & 0.325 & [0.902, 1.000] & 74 & 15 \\
 & new (article only) & 1.000 & 0.001 & [1.000, 1.000] & 74 & 2 \\
\bottomrule
\end{tabular}
\caption{Pangram scores before humanization by domain (2013 to 2015).}
\label{tab:pangram_domain_pre_2015}
\end{table*}

\begin{table*}[h]
\centering
\scriptsize
\begin{tabular}{lllrrrr}
\toprule
Domain & Condition & Mean & SD & [P25, P75] & $n$ & Outliers \\
\midrule
Chemistry & original & 0.026 & 0.146 & [0.000, 0.000] & 95 & 16 \\
 & refine (abs.\ only) & 0.777 & 0.360 & [0.482, 1.000] & 95 & 0 \\
 & refine (abs.+paper) & 0.820 & 0.341 & [0.863, 1.000] & 95 & 18 \\
 & new (article only) & 0.941 & 0.215 & [1.000, 1.000] & 95 & 12 \\
\addlinespace
Computer science & original & 0.099 & 0.289 & [0.000, 0.000] & 83 & 14 \\
 & refine (abs.\ only) & 0.736 & 0.392 & [0.426, 1.000] & 83 & 0 \\
 & refine (abs.+paper) & 0.704 & 0.399 & [0.339, 1.000] & 83 & 0 \\
 & new (article only) & 0.964 & 0.165 & [1.000, 1.000] & 83 & 13 \\
\addlinespace
Political science & original & 0.245 & 0.428 & [0.000, 0.168] & 79 & 19 \\
 & refine (abs.\ only) & 0.877 & 0.273 & [0.980, 1.000] & 79 & 17 \\
 & refine (abs.+paper) & 0.892 & 0.266 & [1.000, 1.000] & 79 & 17 \\
 & new (article only) & 1.000 & 0.002 & [1.000, 1.000] & 79 & 2 \\
\addlinespace
Theology & original & 0.280 & 0.433 & [0.000, 0.838] & 79 & 0 \\
 & refine (abs.\ only) & 0.933 & 0.214 & [0.998, 1.000] & 79 & 17 \\
 & refine (abs.+paper) & 0.956 & 0.196 & [1.000, 1.000] & 79 & 5 \\
 & new (article only) & 0.999 & 0.010 & [1.000, 1.000] & 79 & 2 \\
\bottomrule
\end{tabular}
\caption{Pangram scores before humanization by domain (2023 to 2025).}
\label{tab:pangram_domain_pre_2025}
\end{table*}

\textbf{Pooled and domain summaries after humanization:}
After humanization, pooled means fall to $\approx 0.03$ (Table~\ref{tab:pangram_ranges_post}), consistent with near-universal false negatives in Table~\ref{tab:error_rates}. Residual variance remains in non-STEM domains: political science and theology still show non-zero SD and outliers on some conditions (Tables~\ref{tab:pangram_domain_post_2015} and~\ref{tab:pangram_domain_post_2025}), indicating incomplete score collapse for a subset of abstracts rather than uniform evasion.

\begin{table*}[h]
\centering
\small
\begin{tabular}{lllrrrr}
\toprule
Period & Condition & Mean & SD & [P25, P75] & $n$ & Outliers \\
\midrule
2013 to 2015 & original & 0.003 & 0.057 & [0.000, 0.000] & 306 & 1 \\
2013 to 2015 & refine (abs.\ only) & 0.033 & 0.178 & [0.000, 0.000] & 306 & 10 \\
2013 to 2015 & refine (abs.+paper) & 0.013 & 0.114 & [0.000, 0.000] & 306 & 4 \\
2013 to 2015 & new (article only) & 0.000 & 0.000 & [0.000, 0.000] & 306 & 0 \\
2023 to 2025 & original & 0.033 & 0.178 & [0.000, 0.000] & 336 & 11 \\
2023 to 2025 & refine (abs.\ only) & 0.036 & 0.186 & [0.000, 0.000] & 336 & 13 \\
2023 to 2025 & refine (abs.+paper) & 0.033 & 0.178 & [0.000, 0.000] & 336 & 11 \\
2023 to 2025 & new (article only) & 0.033 & 0.178 & [0.000, 0.000] & 336 & 11 \\
\bottomrule
\end{tabular}
\caption{Pangram score summaries after humanization, pooled across domains.}
\label{tab:pangram_ranges_post}
\end{table*}

\begin{table*}[h]
\centering
\scriptsize
\begin{tabular}{lllrrrr}
\toprule
Domain & Condition & Mean & SD & [P25, P75] & $n$ & Outliers \\
\midrule
Chemistry & original & 0.000 & 0.000 & [0.000, 0.000] & 84 & 0 \\
 & refine (abs.\ only) & 0.000 & 0.000 & [0.000, 0.000] & 84 & 0 \\
 & refine (abs.+paper) & 0.000 & 0.000 & [0.000, 0.000] & 84 & 0 \\
 & new (article only) & 0.000 & 0.000 & [0.000, 0.000] & 84 & 0 \\
\addlinespace
Computer science & original & 0.014 & 0.120 & [0.000, 0.000] & 70 & 1 \\
 & refine (abs.\ only) & 0.014 & 0.120 & [0.000, 0.000] & 70 & 1 \\
 & refine (abs.+paper) & 0.000 & 0.000 & [0.000, 0.000] & 70 & 0 \\
 & new (article only) & 0.000 & 0.000 & [0.000, 0.000] & 70 & 0 \\
\addlinespace
Political science & original & 0.000 & 0.000 & [0.000, 0.000] & 78 & 0 \\
 & refine (abs.\ only) & 0.064 & 0.247 & [0.000, 0.000] & 78 & 5 \\
 & refine (abs.+paper) & 0.026 & 0.159 & [0.000, 0.000] & 78 & 2 \\
 & new (article only) & 0.000 & 0.000 & [0.000, 0.000] & 78 & 0 \\
\addlinespace
Theology & original & 0.000 & 0.000 & [0.000, 0.000] & 74 & 0 \\
 & refine (abs.\ only) & 0.054 & 0.228 & [0.000, 0.000] & 74 & 4 \\
 & refine (abs.+paper) & 0.027 & 0.163 & [0.000, 0.000] & 74 & 2 \\
 & new (article only) & 0.000 & 0.000 & [0.000, 0.000] & 74 & 0 \\
\bottomrule
\end{tabular}
\caption{Pangram scores after humanization by domain (2013 to 2015).}
\label{tab:pangram_domain_post_2015}
\end{table*}

\begin{table*}[h]
\centering
\scriptsize
\begin{tabular}{lllrrrr}
\toprule
Domain & Condition & Mean & SD & [P25, P75] & $n$ & Outliers \\
\midrule
Chemistry & original & 0.000 & 0.000 & [0.000, 0.000] & 95 & 0 \\
 & refine (abs.\ only) & 0.000 & 0.000 & [0.000, 0.000] & 95 & 0 \\
 & refine (abs.+paper) & 0.000 & 0.000 & [0.000, 0.000] & 95 & 0 \\
 & new (article only) & 0.000 & 0.000 & [0.000, 0.000] & 95 & 0 \\
\addlinespace
Computer science & original & 0.024 & 0.154 & [0.000, 0.000] & 83 & 2 \\
 & refine (abs.\ only) & 0.012 & 0.110 & [0.000, 0.000] & 83 & 1 \\
 & refine (abs.+paper) & 0.024 & 0.154 & [0.000, 0.000] & 83 & 2 \\
 & new (article only) & 0.000 & 0.000 & [0.000, 0.000] & 83 & 0 \\
\addlinespace
Political science & original & 0.076 & 0.267 & [0.000, 0.000] & 79 & 6 \\
 & refine (abs.\ only) & 0.063 & 0.245 & [0.000, 0.000] & 79 & 5 \\
 & refine (abs.+paper) & 0.076 & 0.267 & [0.000, 0.000] & 79 & 6 \\
 & new (article only) & 0.076 & 0.267 & [0.000, 0.000] & 79 & 6 \\
\addlinespace
Theology & original & 0.038 & 0.192 & [0.000, 0.000] & 79 & 3 \\
 & refine (abs.\ only) & 0.078 & 0.267 & [0.000, 0.000] & 79 & 7 \\
 & refine (abs.+paper) & 0.038 & 0.192 & [0.000, 0.000] & 79 & 3 \\
 & new (article only) & 0.063 & 0.245 & [0.000, 0.000] & 79 & 5 \\
\bottomrule
\end{tabular}
\caption{Pangram scores after humanization by domain (2023 to 2025).}
\label{tab:pangram_domain_post_2025}
\end{table*}

\section{Full text-feature correlation results}
\label{app:text_feature_correlations}

Tables~\ref{tab:textcorr_2025} and~\ref{tab:textcorr_2015} report the full pre-humanization text-feature correlation results. Correlations are Spearman's $\rho$ between feature ratios and detector AI score. The domain-adjusted column reports correlation after centering within domain. $q$ is the Benjamini-Hochberg FDR across the 10 tests per collection.

\begin{table}[h]
\centering
\footnotesize
\setlength{\tabcolsep}{3pt}
\resizebox{\columnwidth}{!}{%
\begin{tabular}{llrrrr}
\toprule
Feature & Detector & $\rho$ & $p$ & $q$ & Domain-adj.\ $r$ \\
\midrule
Numeric token ratio & Pangram & -0.146 & $<0.001$ & $<0.001$ & -0.148 \\
Numeric token ratio & GPTZero & -0.171 & $<0.001$ & $<0.001$ & -0.098 \\
Non-alphabetic token ratio & Pangram & -0.149 & $<0.001$ & $<0.001$ & -0.085 \\
Non-alphabetic token ratio & GPTZero & -0.197 & $<0.001$ & $<0.001$ & -0.109 \\
Long token ratio & Pangram & 0.336 & $<0.001$ & $<0.001$ & 0.421 \\
Long token ratio & GPTZero & 0.304 & $<0.001$ & $<0.001$ & 0.321 \\
Acronym ratio & Pangram & -0.055 & 0.044 & 0.044 & -0.047 \\
Acronym ratio & GPTZero & -0.082 & 0.004 & 0.004 & -0.024 \\
AWL token ratio & Pangram & 0.346 & $<0.001$ & $<0.001$ & 0.412 \\
AWL token ratio & GPTZero & 0.320 & $<0.001$ & $<0.001$ & 0.290 \\
\bottomrule
\end{tabular}%
}
\caption{Text-feature correlations (2023 to 2025 collection; $n=1{,}344$).}
\label{tab:textcorr_2025}
\end{table}

\begin{table}[h]
\centering
\footnotesize
\setlength{\tabcolsep}{3pt}
\resizebox{\columnwidth}{!}{%
\begin{tabular}{llrrrr}
\toprule
Feature & Detector & $\rho$ & $p$ & $q$ & Domain-adj.\ $r$ \\
\midrule
Numeric token ratio & Pangram & -0.057 & 0.045 & 0.045 & -0.120 \\
Numeric token ratio & GPTZero & -0.074 & 0.009 & 0.009 & -0.059 \\
Non-alphabetic token ratio & Pangram & 0.005 & 0.869 & 0.869 & -0.076 \\
Non-alphabetic token ratio & GPTZero & -0.119 & $<0.001$ & $<0.001$ & -0.045 \\
Long token ratio & Pangram & 0.355 & $<0.001$ & $<0.001$ & 0.366 \\
Long token ratio & GPTZero & 0.287 & $<0.001$ & $<0.001$ & 0.304 \\
Acronym ratio & Pangram & -0.004 & 0.887 & 0.887 & 0.002 \\
Acronym ratio & GPTZero & -0.035 & 0.199 & 0.199 & -0.007 \\
AWL token ratio & Pangram & 0.331 & $<0.001$ & $<0.001$ & 0.400 \\
AWL token ratio & GPTZero & 0.258 & $<0.001$ & $<0.001$ & 0.277 \\
\bottomrule
\end{tabular}%
}
\caption{Text-feature correlations (2013 to 2015 collection; $n=1{,}224$ Pangram-scored texts with features).}
\label{tab:textcorr_2015}
\end{table}

\section{Length-partial correlations}
\label{app:length_partial}

Table~\ref{tab:length_partial_2025} reports Spearman $\rho$ between detector scores and surface features after residualizing both variables on $\log(1+\text{word count})$ (2023 to 2025, pre-humanization).

\begin{table}[h]
\centering
\small
\begin{tabular}{llrr}
\toprule
Feature & Detector & Raw $\rho$ & Partial $\rho$ \\
\midrule
AWL token ratio & Pangram & 0.346 & 0.337 \\
AWL token ratio & GPTZero & 0.320 & 0.261 \\
Long token ratio & Pangram & 0.336 & 0.321 \\
Long token ratio & GPTZero & 0.304 & 0.276 \\
Numeric token ratio & Pangram & $-0.146$ & $-0.132$ \\
Numeric token ratio & GPTZero & $-0.171$ & $-0.134$ \\
\bottomrule
\end{tabular}
\caption{Length-partial Spearman $\rho$ (2023 to 2025, pre-humanization).}
\label{tab:length_partial_2025}
\end{table}

\section{Detector score distributions}
\label{app:score_distributions}

Table~\ref{tab:error_rates} reports error rates at a single threshold; the figures below show the underlying score distributions in the 2023 to 2025 collection.

\textbf{Pre-humanization separation (RQ1).}
Figure~\ref{fig:pangram_pre_2025} plots Pangram scores by domain and rewrite condition before humanization. This figure is central to RQ1 because it makes the assisted-writing risk visible as a distributional shift: \emph{original} abstracts cluster near zero, while even \emph{refine (abstract only)} edits move substantial mass toward high scores - the pattern behind proxy false-positive risk on light editing. It also shows that \emph{new (article only)} scores sit near certainty, explaining why false-negative rates on full-synthesis rewrites are near zero before humanization.

\begin{figure*}[h]
    \centering
    \includegraphics[width=\linewidth]{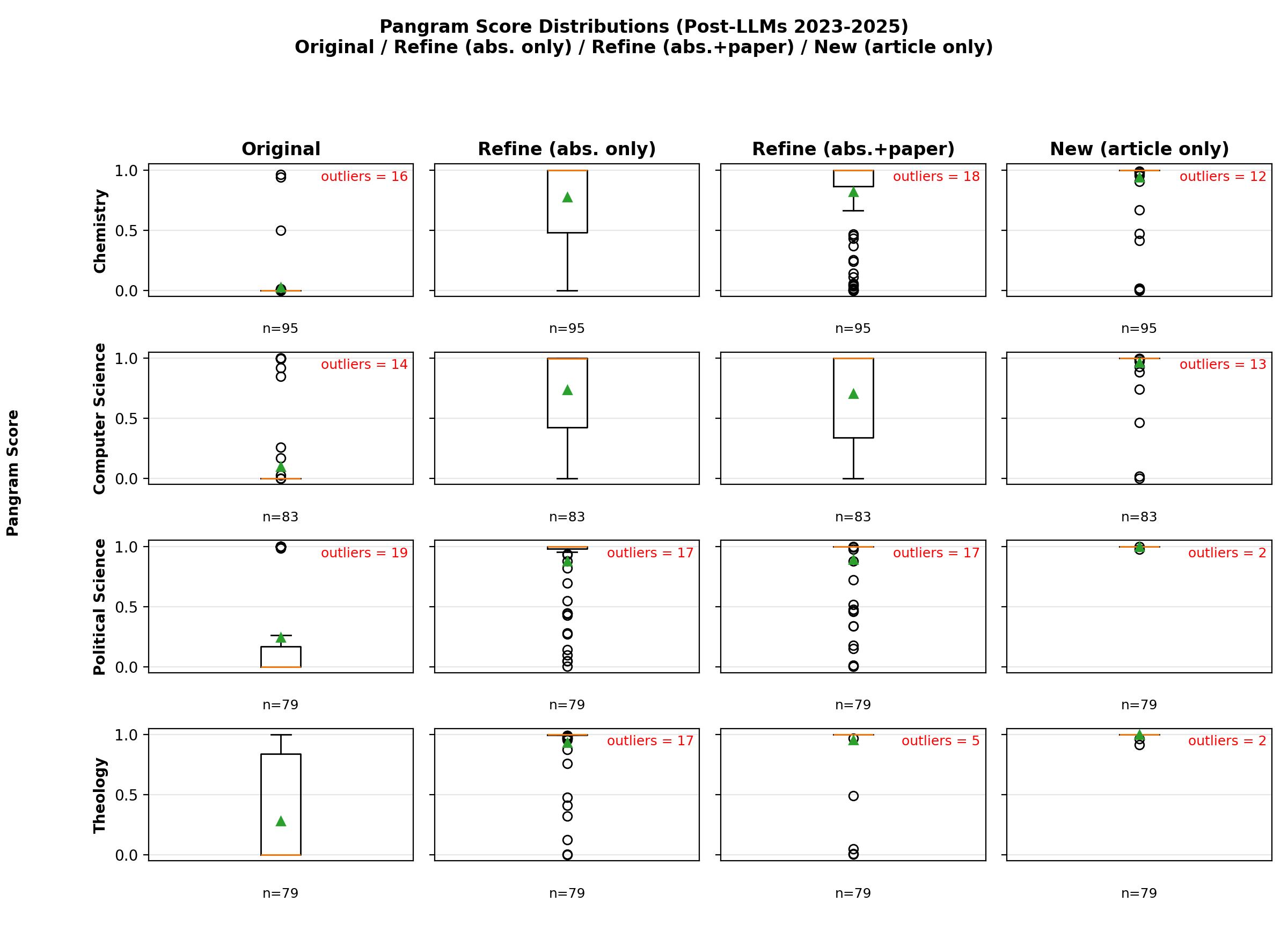}
    \caption{Pangram scores before humanization (2023 to 2025), by domain (rows) and rewrite condition (columns). Originals cluster low; \emph{refine (abstract only)} and heavier rewrites shift right.}
    \label{fig:pangram_pre_2025}
\end{figure*}

\textbf{Post-humanization collapse (RQ3).}
Figure~\ref{fig:pangram_post_2025} uses the same layout after Undetectable humanization. Read together with Figure~\ref{fig:pangram_pre_2025}, it is the main visual evidence for RQ3: AI-labeled variants that scored near 1.0 pre-humanization collapse toward 0.0 post-humanization, which drives the $>96\%$ false-negative rates in Table~\ref{tab:error_rates} and shows that commercial evasion can erase detector signal at scale.

\begin{figure*}[h]
    \centering
    \includegraphics[width=\linewidth]{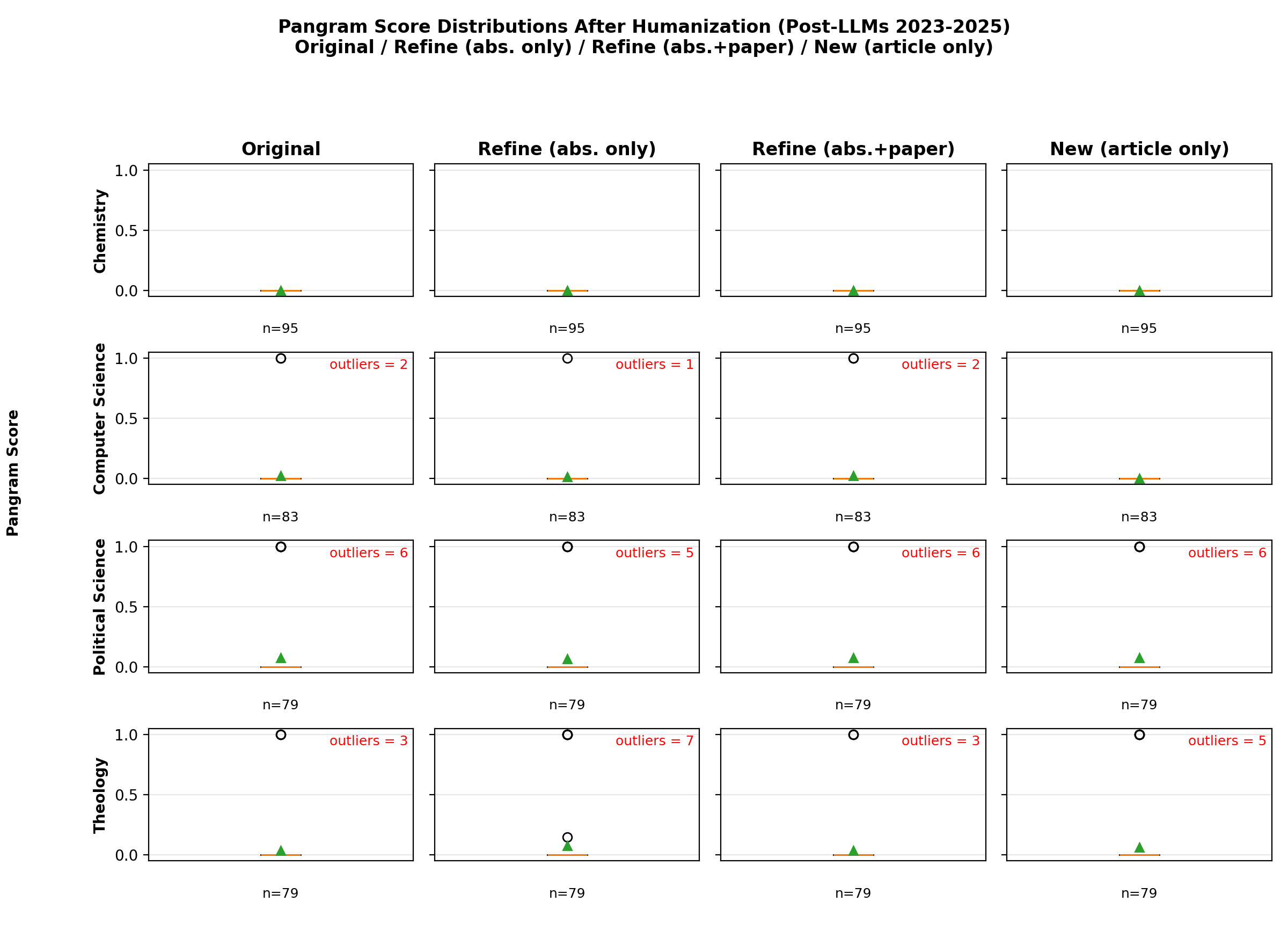}
    \caption{Pangram scores after humanization (2023 to 2025). Compare to Figure~\ref{fig:pangram_pre_2025}: high pre-humanization scores on AI-labeled variants largely disappear.}
    \label{fig:pangram_post_2025}
\end{figure*}

\FloatBarrier
\section{Robustness and threshold sensitivity}
\label{app:robustness}

This section complements Table~\ref{tab:error_rates} by asking how detector behavior changes with the scoring threshold $\tau$ and whether standard precision-recall (PR) summaries align with academic-integrity policy. ROC and PR curves in the main text and in Figure~\ref{fig:rewrite_condition_roc_pr} use \textbf{2013 to 2015} pre-humanization scores so that \emph{original} abstracts can serve as negatives with interpretable proxy FPR. Figures~\ref{fig:assisted_editing_roc_pr_recent} and~\ref{fig:rewrite_condition_roc_pr_recent} repeat the same tasks on 2023 to 2025, where original-axis rates are flag rates (unobserved real-world AI use). Threshold trade-off plots use 2023 to 2025 throughout.

\subsection{Why we do not report pooled precision-recall}
A common evaluation pools \emph{original} abstracts as negatives and \emph{all} LLM rewrites (refine, refine with paper, and new) as positives. Under that label scheme, both Pangram and GPTZero achieve AUC-PR $\approx 0.96$ because \emph{new (article only)} outputs score near 1.0 and dominate the positive class. That number suggests excellent separability but answers the wrong question for integrity offices: policy concern is whether a detector can distinguish \textbf{human drafts from light AI editing}, not whether it can detect full-synthesis rewrites that already saturate the score scale. We therefore report PR only for the assisted-editing subset below, plus a threshold trade-off plot that tracks three rates institutions actually care about.

\subsection{Assisted-editing ROC and precision-recall}
We restrict to paired rows on the same paper: \emph{original} abstract $=$ negative (proxy human), \emph{refine (abstract only)} $=$ positive (proxy light AI edit on human-source text). Scores are detector outputs in $[0,1]$; a text is flagged when $s \ge \tau$. Figure~\ref{fig:assisted_editing_roc} in the main text shows ROC only (2013 to 2015); Figure~\ref{fig:assisted_editing_roc_pr_recent} shows the matching 2023 to 2025 ROC/PR panel. \emph{Original} is not its own curve. It defines the false-positive rate axis.

\subsection{ROC and PR by rewrite condition (2013 to 2015)}
Figure~\ref{fig:rewrite_condition_roc_pr} repeats the analysis for each rewrite type against the same pre-LLM-era originals: \emph{refine (abstract only)}, \emph{refine (abstract + article)}, and \emph{new (article only)}, plus a dashed \emph{pooled AI} curve (all three LLM conditions as positives). AUC-PR is high ($0.93$-$1.00$) because Gemini rewrites separate cleanly from unflagged originals at $\tau=0.50$; this is a best-case separability benchmark, not current flagging exposure.

\begin{figure*}[h]
    \centering
    \includegraphics[width=\linewidth]{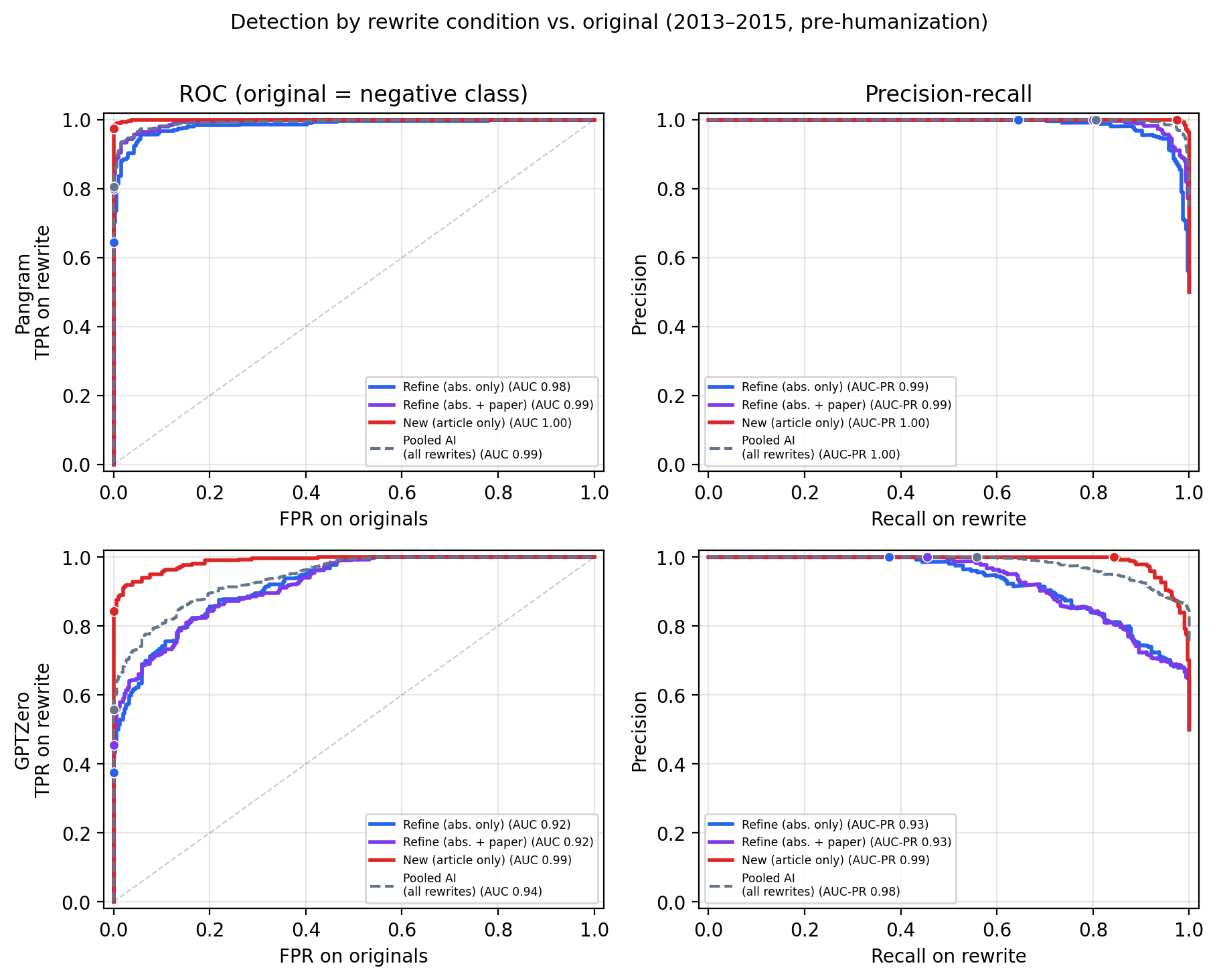}
    \caption{ROC and precision-recall by rewrite condition vs.\ \emph{original} (2013 to 2015, pre-humanization). Four curves per panel: three single-condition tasks (solid) and pooled all LLM rewrites (dashed). Rows: Pangram and GPTZero. Markers: $\tau=0.50$.}
    \label{fig:rewrite_condition_roc_pr}
\end{figure*}

\subsection{2023 to 2025 replication (flag rate on originals)}
On recent abstracts, positives on the original axis are \textbf{flag rates}, not confirmed false positives. Pangram AUC-PR on assisted editing falls to $0.88$; at $\tau=0.50$ it flags 80\% of light edits with precision $0.84$ while 15\% of originals are also flagged. GPTZero AUC-PR is $0.85$ with 49\% refine recall. Pooled-AI AUC-PR remains $\approx 0.96$ (Pangram) and $0.95$ (GPTZero); the assisted-editing panel is in Figure~\ref{fig:assisted_editing_roc_pr_recent} (main text); per-condition curves are in Figure~\ref{fig:rewrite_condition_roc_pr_recent}.

\begin{figure*}[h]
    \centering
    \includegraphics[width=\linewidth]{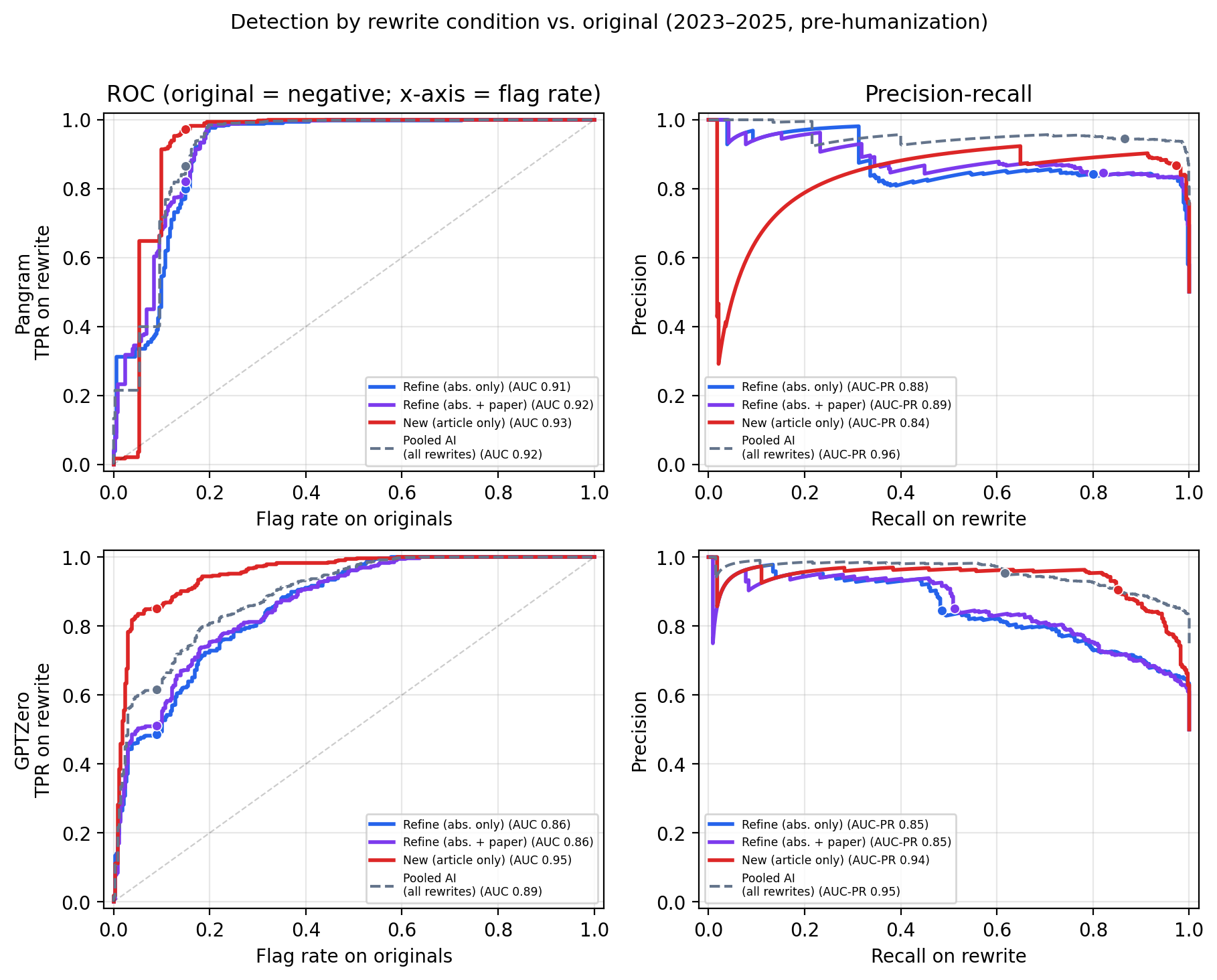}
    \caption{Rewrite-condition ROC/PR on 2023 to 2025 (pre-humanization); same layout as Figure~\ref{fig:rewrite_condition_roc_pr}.}
    \label{fig:rewrite_condition_roc_pr_recent}
\end{figure*}

\subsection{Threshold trade-off curves}
For each detector and each $\tau \in \{0.3,0.4,0.5,0.6,0.7\}$, we report three percentages on the 2023 to 2025 collection:
\begin{enumerate}
  \item \textbf{Original flag rate} (gray): fraction of unmodified \emph{original} abstracts with $s \ge \tau$. On recent prose this mixes proxy false positives with unobserved real AI use; it still bounds how often human-labeled text is flagged.
  \item \textbf{Refine (abstract only) flagged rate} (solid blue/red): fraction of light edits flagged-the assisted-writing exposure from RQ1.
  \item \textbf{FNR on AI-labeled pool} (dashed orange): fraction of pooled LLM rewrites (all three conditions) scoring \emph{below} $\tau$-missed AI under proxy labels.
\end{enumerate}

There is no $\tau$ that simultaneously keeps all three rates low. For Pangram, original flag rate and refine-flag rate stay near 15\% and 80\% respectively across $\tau \in [0.4,0.6]$; raising $\tau$ from 0.4 to 0.5 increases missed AI-labeled rewrites (FNR 10.5\% to 13.5\%) and does not make light edits safe from flags. For GPTZero, refine-flag rate is already near 50\% at $\tau=0.50$ while FNR on the AI pool remains 38\%; lowering $\tau$ increases flags on originals without catching most light edits. The vertical dotted line marks our main-text operating point $\tau=0.50$.

The table reports the same three quantities at $\tau \in \{0.4,0.5,0.6\}$; the figure extends the sweep to 0.3 and 0.7 and places Pangram and GPTZero side by side.

\subsection{Tabulated threshold sensitivity}
Table~\ref{tab:threshold_sensitivity} in the main text reports original flag rates and AI-labeled false-negative rates for $\tau \in \{0.4,0.5,0.6\}$ in the 2023 to 2025 collection (pre-humanization). Figure~\ref{fig:threshold_tradeoff} extends the sweep to $\tau \in \{0.3,\ldots,0.7\}$.

\section{Automatically selected examples}
\label{app:examples}

We provide two \emph{automatically selected} examples illustrating extreme false-negative transitions after humanization (pre scores near 1.0, post scores near 0.0) without manual cherry-picking. For each time window (2013 to 2015 and 2023 to 2025), we selected one paper-variant pair \textbf{deterministically} from items with complete pre/post outputs. Specifically, we (i) restricted to pairs with both Pangram and GPTZero available when possible, (ii) computed the mean detector score change after humanization \((\Delta = \mathrm{post}-\mathrm{pre})\) across the available detectors for each paper-variant pair, and (iii) selected the pair with the \textbf{largest drop} (most negative \(\Delta\)). This selection rule is fully specified in our analysis script and does not use randomness.

The 2023 to 2025 auto-selected example (paper ID \texttt{W4401907927}, political science, \emph{new (article only)}) shows a complete detector flip: Pangram changed from $1.000$ to $0.000$ and GPTZero changed from $1.000$ to $0.000$ after humanization (mean $\Delta=-1.000$ across detectors). The 2013 to 2015 auto-selected example (\texttt{W3009617460}, political science, \emph{refine (abstract + article)}) shows the same pattern.

To diagnose why the flip occurs, we compared measurable pre/post text properties for the selected examples. In the 2023 to 2025 case, long-token ratio dropped from $0.260$ to $0.100$, Academic Word List (AWL) token ratio dropped from $0.252$ to $0.006$, and lexical diversity (type-token ratio) dropped from $0.771$ to $0.502$, while the rewrite became much longer (129 to 311 words) and shifted to much longer sentence units (average 25.8 to 62.2 words per sentence). In the 2013 to 2015 case, long-token ratio also decreased ($0.082 \rightarrow 0.054$) and lexical diversity decreased ($0.629 \rightarrow 0.549$). These changes are directionally consistent with our corpus-level findings that detectors assign higher AI scores to texts with more long technical tokens, and lower scores to texts that reduce those signals. In short, the humanizer appears to suppress detector-salient cues rather than preserve the original stylistic profile.

\section{Linguistic mechanisms of humanization}
\label{app:humanization_linguistics}

Undetectable AI v11 is a proprietary service; we treat it as a black box but measure how its outputs differ linguistically from the inputs sent for humanization. For every paper-variant pair with both \texttt{original\_abstract} and \texttt{humanized\_abstract} under \texttt{humanization/} ($n=2{,}568$ pairs across both time collections), we computed the surface features defined in Section~\ref{sec:error_operationalization}, plus the fraction of sentences with at most eight words (\emph{short-sentence ratio}) as a simple fragmentation proxy. We tested paired shifts (post minus pre) with two-sided permutation tests (5{,}000 iterations; seed 42) and correlated feature deltas with Pangram and GPTZero score deltas on matched rows.

\textbf{Systematic suppression of detector-salient cues.}
Table~\ref{tab:humanization_linguistics} in the main text summarizes key pooled shifts; Table~\ref{tab:humanization_linguistics_full} below lists the full feature set. Humanization \emph{reduced} long-token ratio (0.184 to 0.138; mean $\Delta=-0.045$; 88.2\% of pairs decreased; $p<0.001$), AWL token ratio (0.157 to 0.121; $\Delta=-0.036$; 82.9\% decreased; $p<0.001$), and type-token ratio (0.700 to 0.605; $\Delta=-0.094$; 93.0\% decreased; $p<0.001$). Numeric token ratio did not shift materially ($p=0.83$). These directions match our pre-humanization finding that higher Pangram and GPTZero scores correlate with greater long-token and AWL density: the humanizer moves text along the same axes that separate high-scoring from low-scoring abstracts.

\textbf{Length and syntax.}
Humanization also \emph{increased} length: mean word count rose from 180 to 229 ($\Delta \approx +49$ words; $p<0.001$) and average words per sentence from 23.6 to 28.4 ($\Delta \approx +4.8$; $p<0.001$). Short-sentence ratio increased only slightly (0.063 to 0.072; $p<0.001$), so evasion is not primarily driven by chopping prose into many brief sentences. Instead, outputs tend to be longer and more sententially expanded while using fewer long technical tokens and fewer academic-wordlist hits.

\textbf{Coupling to detector score drops.}
When a pair's long-token, AWL, or type-token ratios fell more sharply, Pangram and GPTZero scores tended to fall more sharply as well (Spearman $\rho \approx 0.30$ to $0.40$ on $\Delta$ features vs.\ $\Delta$ scores; $p<0.001$ for all three features on both detectors). Average words-per-sentence increases were weakly associated with smaller score drops on Pangram ($\rho=-0.12$) and not reliably on GPTZero ($\rho=-0.05$). Together, the corpus-level shifts and the score-feature couplings suggest a mechanistic story: Undetectable humanization neutralizes commercial detectors largely by diluting the surface academic and technical token patterns those detectors already treat as AI-like, not by uniformly fragmenting syntax.

\begin{table*}[h]
\centering
\small
\begin{tabular}{lrrrrr}
\toprule
Feature & Pre & Post & $\Delta$ & \% decreased & $p$ \\
\midrule
Long-token ratio & 0.184 & 0.138 & $-0.045$ & 88.2\% & $<0.001$ \\
AWL token ratio & 0.157 & 0.121 & $-0.036$ & 82.9\% & $<0.001$ \\
Type-token ratio & 0.700 & 0.605 & $-0.094$ & 93.0\% & $<0.001$ \\
Numeric token ratio & 0.017 & 0.017 & $0.000$ & 38.9\% & 0.833 \\
Non-alphabetic token ratio & 0.079 & 0.068 & $-0.011$ & 75.2\% & $<0.001$ \\
Avg.\ words per sentence & 23.6 & 28.4 & $+4.8$ & 22.2\% & $<0.001$ \\
Short-sentence ratio ($\leq 8$ words) & 0.063 & 0.072 & $+0.009$ & 16.1\% & $<0.001$ \\
Word count & 180 & 229 & $+49$ & 10.2\% & $<0.001$ \\
\bottomrule
\end{tabular}
\caption{Full pre/post linguistic features after Undetectable AI v11 humanization (pooled corpus; $n=2{,}568$ paper-variant pairs). $\Delta$ is post minus pre. $p$ is a two-sided permutation test on paired $\Delta$.}
\label{tab:humanization_linguistics_full}
\end{table*}

\section{Cost and scope of the pipeline}
\label{app:cost}
Our design trades breadth across rewrite \emph{intensity} and detectors for feasibility at corpus scale. We sampled 800 English abstracts (642 after length and PDF filtering; Section~\ref{sec:corpus_pipeline}). Each retained paper contributes four scored texts: one \emph{original} and three LLM outputs (\emph{refine (abstract only)}, \emph{refine (abstract + article)}, \emph{new (article only)}).

\textbf{Why one rewrite model:}
We used a single generator (Gemini 3 Flash via OpenRouter) rather than a multi-model sweep. Adding $K$ rewrite models would multiply LLM inference cost by $K$ while holding our error-analysis questions fixed (proxy labels, detector comparison, humanization). The factorial structure we need is rewrite \emph{condition} and detector, not vendor-specific rewrite style.

\textbf{Order-of-magnitude call accounting:}
Let $N$ denote retained papers ($N = 642$). Per paper:
\begin{itemize}
  \item \textbf{LLM rewrites:} $3$ calls (one per rewrite condition; the original is not regenerated).
  \item \textbf{Pre-humanization detection:} $4$ variants $\times$ $3$ detectors (Pangram, GPTZero, LLM-assisted baseline) $= 12$ scored texts per paper.
  \item \textbf{Humanization:} $4$ variants $\times$ $1$ Undetectable pass (paid credits per abstract; not repeated per detector).
  \item \textbf{Post-humanization detection:} $4$ variants $\times$ $3$ detectors (Pangram, GPTZero, and LLM-assisted baseline).
\end{itemize}
Aggregating over the corpus, pre-detection scales as $N \times 4 \times 3$, humanization as $N \times 4$, and post-detection as $N \times 4 \times 3$.

\textbf{LLM-assisted baseline role:}
The LLM-assisted baseline flags a high share of recent \emph{original} abstracts (Table~\ref{tab:error_rates}) and is treated as supplementary. After humanization, Pangram and GPTZero scores collapse toward zero for AI-labeled variants, but the LLM-assisted baseline still flags most humanized rewrites (post-humanization FNR 30.5\% and 19.7\%). Main-text evasion claims, therefore, focus on commercial detectors (RQ3).

\fi
\end{document}